\documentclass[11pt,letterpaper]{article}

\usepackage{booktabs}

\newcommand\spacingset[1]{\renewcommand{\baselinestretch}%
  {#1}\small\normalsize}

\usepackage{latexsym}
\usepackage{amssymb,amsmath, bm,pgfplots,tikz,bbm}
\usepackage{graphicx}
\usepackage{marvosym}
\usepackage{multirow,float}
\usepackage{algpseudocode}
\usepackage{caption}
\usepackage{mathtools}
\usepackage{subcaption}
\usepackage{comment}
\usepackage{enumitem}

\usepackage{natbib}
\usepackage{color}
\usepackage[bookmarksopen=true, bookmarksnumbered=true,
pdfstartview=FitH, breaklinks=true, urlbordercolor={0 1 0}, citebordercolor={0 0 1}]{hyperref}

\usepackage{dcolumn}
\newcolumntype{.}{D{.}{.}{-1}}
\newcolumntype{d}[1]{D{.}{.}{#1}}

\usepackage{theorem}
\theoremstyle{plain}
\theoremheaderfont{\scshape}
\newtheorem{theorem}{Theorem}
\newtheorem{proposition}{Proposition}
\newtheorem{assumption}{Assumption}

\newcommand{\qed}{\hfill \ensuremath{\Box}}
\newcommand{\indep}{\mbox{$\perp\!\!\!\perp$}}

\DeclareMathOperator*{\argmin}{argmin}

\newcommand{\norm}[1]{\left\lVert#1\right\rVert}
\newenvironment{proof}{\vspace{1ex}\noindent{\bf Proof}\hspace{0.5em}}
{\hfill\qed\vspace{1ex}}
\usepackage{kantlipsum}
\allowdisplaybreaks

\graphicspath{{Manuscripts_application/figs/}}

\usepackage{rotating}

\usepackage{arydshln}
\usepackage{threeparttable}

\usepackage[compact]{titlesec}

\usepackage[ruled,linesnumbered,vlined]{algorithm2e}
\usepackage{pifont}

\newtheorem{definition}{Definition}

\newcommand\E{\mathbb{E}}
\newcommand\R{\mathbb{R}}
\newcommand\cR{\mathcal{R}}
\renewcommand\P{\mathbb{P}}

\newcommand\cQ{\mathcal{Q}}
\newcommand\bq{\bm{q}}

\newcommand\bP{\bm{P}}

\newcommand\br{\bm{r}}
\newcommand\bR{\bm{R}}

\newcommand\bh{\bm{h}}

\newcommand\bx{\bm{x}}
\newcommand\bX{\bm{X}}

\newcommand\boldf{\bm{f}}
\newcommand\bg{\bm{g}}
\newcommand\bu{\bm{u}}
\newcommand\bU{\bm{U}}
\newcommand\cD{\mathcal{D}}
\newcommand\cT{\mathcal{T}}
\newcommand\cU{\mathcal{U}}

\newcommand\cX{\mathcal{X}}
\newcommand\cY{\mathcal{Y}}

\newcommand\bgamma{\boldsymbol{\gamma}}
\newcommand\btheta{\boldsymbol{\theta}}

\newcommand\blambda{\boldsymbol{\lambda}}
\usetikzlibrary{decorations.markings}
\usetikzlibrary{decorations.pathmorphing}
\usetikzlibrary{shapes.geometric, arrows}
\usetikzlibrary{arrows,decorations.pathmorphing,backgrounds,positioning,fit,matrix}
\usetikzlibrary{shapes,decorations,arrows,calc,arrows.meta,fit,positioning}
\tikzset{auto,node distance =1 cm and 1 cm,semithick,
	state/.style ={circle, draw, minimum width = 0.7 cm},
	point/.style = {circle, draw, inner sep=0.04cm,fill,node contents={}},
	bidirected/.style={Latex-Latex,dashed},
	el/.style = {inner sep=2pt, align=left, sloped}
}

\graphicspath{{figs/}}

\begin{document}

\title{Leveraging Generative Artificial Intelligence for Causal Inference with Unstructured Data\footnote{An open-source Python package,
    {\tt GPI: GenAI-Powered Inference}, that implements the proposed
    methodology is available at \url{https://gpi-pack.github.io/}. We
    thank Naoki Egami for helpful comments.}}
\author{ Kosuke Imai\thanks{Professor, Department of Government and
    Department of Statistics, Harvard University, Cambridge, MA
    02138. Phone: 617--384--6778, Email:
    \href{mailto:Imai@Harvard.Edu}{Imai@Harvard.Edu}, URL:
    \href{https://imai.fas.harvard.edu}{https://imai.fas.harvard.edu}}
  \and Kentaro Nakamura\thanks{Ph.D. student, John F. Kennedy School
    of Government, Harvard University. Email:
    \href{mailto:knakamura@g.harvard.edu}{knakamura@g.harvard.edu}} }
\date{\today}
\maketitle\thispagestyle{empty}

\begin{abstract}
  We introduce GenAI-Powered Inference (GPI), a statistical framework
  for causal inference using unstructured data,
  including text and images. GPI leverages open-source pre-trained Generative
  Artificial Intelligence (GenAI) models---such as large language
  models and diffusion models---not only to generate unstructured data
  at scale but also to extract low-dimensional representations that
  are guaranteed to capture their underlying structure. Applying
  machine learning to these representations, GPI enables estimation of
  causal effects while quantifying estimation
  uncertainty. Unlike existing approaches to representation learning,
  GPI does not require fine-tuning of GenAI models, making it
  computationally efficient and broadly accessible. We illustrate the
  versatility of the GPI framework through three applications: (1)
  estimating the effects of Chinese social media censorship while
  adjusting for textual confounders, (2) isolating the impact of
  specific image features from that of other correlated features in
  the same image, and (3) assessing the persuasiveness of political
  rhetoric. An open-source software package is available for
  implementing GPI.
\end{abstract}


\newpage
\section{Introduction}

The emergence of Generative Artificial Intelligence (GenAI),
including large language models and diffusion models, has transformed
society and scientific research. In this paper, we show that GenAI can
also serve as a powerful tool for causal inference with unstructured
data such as text and images. Specifically, we introduce a new
statistical framework, {\it GenAI-Powered Inference} (GPI), which
leverages open-source GenAI models to improve causal inference with
high-dimensional unstructured data.

We illustrate the power of GPI through the following three applications:
\begin{enumerate}
\item \textbf{Text as confounder}: Estimating the causal effects of a
  treatment variable by adjusting for latent confounders embedded in
  textual data

\item \textbf{Image as Treatment}: Isolating the impact of specific
  image features from that of other correlated features in the same
  image

\item \textbf{Structural model of text}: Fitting a structural model
  that incorporates both observed and latent textual features as
  covariates
\end{enumerate}
A common challenge in these settings is that while high-dimensional
unstructured data are observed, we do not know, {\it a priori}, which
features of such data serve as relevant confounders.

GPI addresses this challenge by leveraging the ability of GenAI to
generate or reproduce unstructured data and extracting
lower-dimensional vector representations that are guaranteed to
contain all the relevant information. Machine learning is then applied
to these representations to estimate a {\it deconfounder}, which is a
summary of confounding information. Because these representations were
used to generate unstructured data, GPI eliminates the need to
explicitly model the data-generating process, fine-tune foundation
models, or estimate text and image representations. This makes GPI a
broadly accessible framework.  Our open-source Python package {\tt
  GPI}, which is available at \url{https://gpi-pack.github.io/},
implements the proposed methodology to further enhance its
applicability in applied research.

\begin{figure}[t]
  \centering \spacingset{1}
  \includegraphics[width=1.0\linewidth]{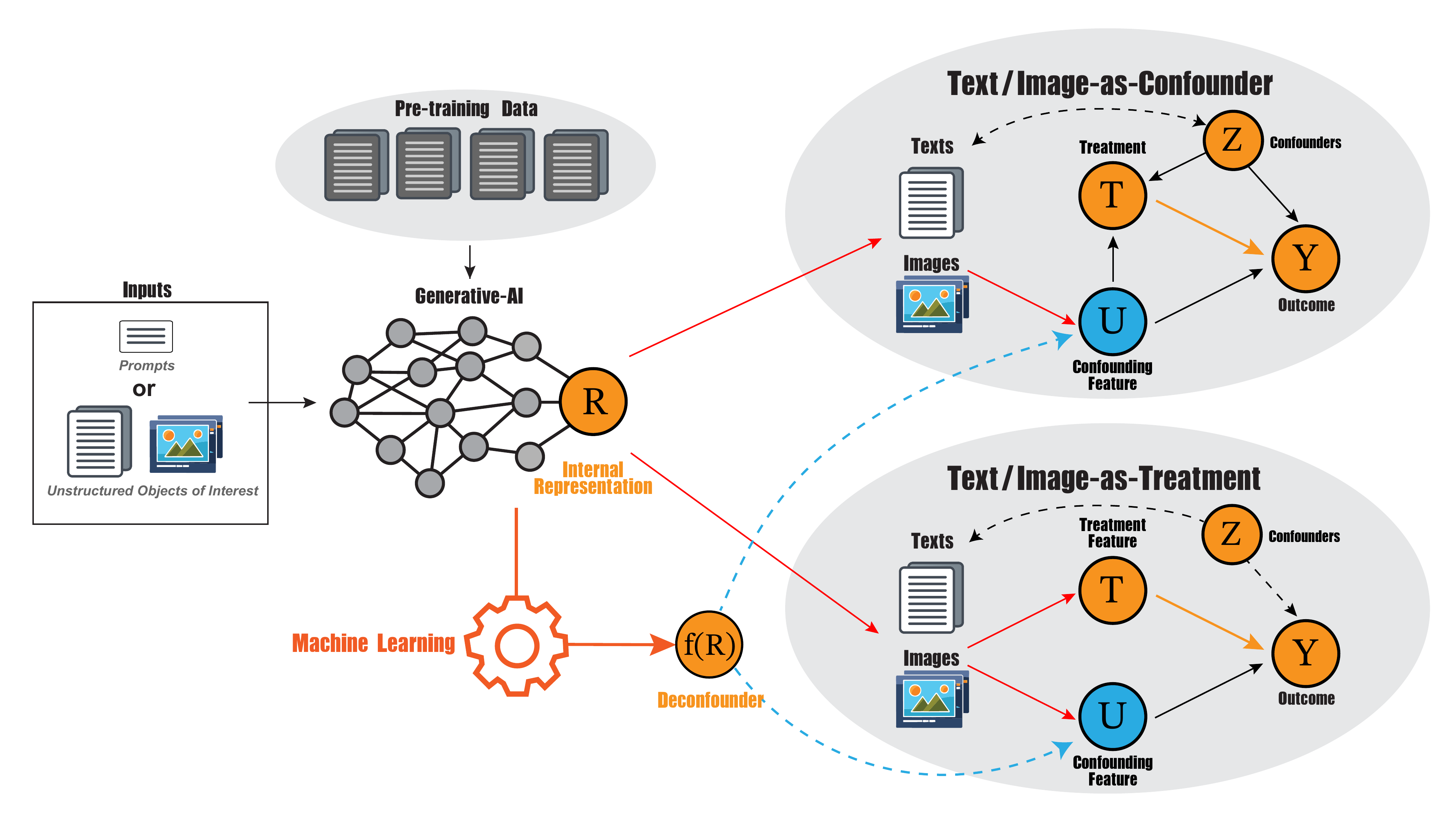}
  \caption{Illustration of GenAI-Powered Inference (GPI).  Inputs to
    generative AI are either prompts or unstructured data (e.g.,
    text/images). Pre-trained models produce internal representations
    $\bR$, which we use for machine learning in two application settings: (i)
    when text/images act as confounders, and (ii) when they serve as
    treatments. In the first setting (top right), the
    treatment-outcome relation is confounded by the latent features of
    unstructured data $\bU$ and observed covariates $\bm Z$. In the
    second setting (bottom right), the treatment depends on
    unstructured data and is similarly confounded. Since $\bU$ is
    unknown, we apply machine learning to estimate a deconfounder
    $\boldf(\bR)$, which is a low-dimensional summary of confounding
    features.  GPI enables unbiased causal inference without modeling
    the generative process or fine-tuning the GenAI model.}
  \label{diagram}
\end{figure}

Before presenting the three empirical applications, we provide a brief
overview of the proposed GPI methodology (see also
\citealt{imai2024causal}). Figure~\ref{diagram} considers two
illustrative settings, in which the goal is to estimate the causal
effect of treatment $T$ on an outcome $Y$. The treatment may represent
some features of unstructured data $\bX$ as in the ``text/image as
treatment'' case (bottom right) or other structured external variables
as in the ``text/image as confounder'' case (top right). In both
settings, the treatment-outcome relationship is confounded by the
latent features $\bU$ of unstructured data $\bX$ as well as other
observed structured confounders $\bm{Z}$.  A key challenge is to
identify these latent (unstructured) confounding features $\bU$ and
adjust for them to estimate causal effects.

To achieve this, GPI leverages GenAI models to generate unstructured
data $\bX$ and extract the corresponding internal representations
$\bR$. For existing text or images, GenAI models are prompted to
reproduce the original content, enabling the recovery of consistent
internal representations. Because we configure GenAI models to produce
$\bX$ as a deterministic function of $\bR$, the latter is guaranteed
to contain all information necessary to generate the former. As a
result, GPI eliminates the need to fine-tune $\bR$, unlike standard
approaches that rely on generic pretrained embeddings such as BERT
(Bidirectional Encoder Representations from Transformers).

The use of $\bR$ offers several important advantages over directly
modeling the raw data $\bX$. First, $\bR$ provides a substantially
lower-dimensional representation of the original high-dimensional
input. Second, it captures rich and relevant information distilled
from the vast data used to train the GenAI model. Third, by relying on
open-source GenAI models that operate deterministically, GPI promotes
scientific replicability. Finally, GPI can further improve statistical
efficiency by combining the results based on multiple GenAI models.

GPI applies machine learning to the internal representation $\bR$ to
estimate a deconfounder $\boldf(\bR)$, which serves as a
low-dimensional approximation of the latent confounding features
$\bU$. This step is critical because directly adjusting for the
high-dimensional unstructured data $\bX$ can violate the positivity
assumption and lead to biased inference \citep{d2021overlap}. While
the deconfounder is not necessarily unique, we formally show that any
valid $\boldf(\bR)$ enables the nonparametric identification of causal
effects when adjusted for together with the observed confounders
$\bm{Z}$.
We propose estimating the deconfounder using a deep neural network
architecture designed to directly encode these necessary properties.

GPI learns a low-dimensional deconfounder $\boldf(\bR)$ that is
predictive of the outcome within each treatment group while remaining
shared across treatment and control groups. By discarding components of $\bR$ that
predict treatment assignment alone, $\boldf(\bR)$ also makes the
overlap assumption more likely to hold than standard approaches in the
literature \citep[e.g.,][]{shi_adapting_2019, veitch_adapting_2020,
  pryzant_causal_2021}.  We emphasize the importance of carefully
tuning the neural network architecture used in GPI: the network must
be sufficiently expressive to preserve outcome-relevant information,
yet parsimonious enough to remove irrelevant variation. In
Appendix~\ref{sec:simulation_studies}, we empirically illustrate this
point through a simple simulation study.  The appendix also provides
theoretical justifications of GPI tailored to each application setting
described below.

\subsection{Related Literature}

A growing body of work addresses causal inference with unstructured
high-dimensional data, often by either estimating low-dimensional
embeddings \citep[e.g.,][]{veitch_adapting_2020, pryzant_causal_2021,
  gui_causal_2023, klaassen_doublemldeep_2024} or modeling the
data-generating process using parametric methods such as topic models
\citep[e.g.,][]{fong_discovery_2016, mozer_matching_2020,
  roberts_adjusting_2020, ahrens_bayesian_2021,
  egami_how_2022}. These approaches typically rely on strong
assumptions, such as the bag-of-words model and fixed embedding
schemes, and face substantial challenges due to the statistical and
computational complexity of high-dimensional unstructured data. This
may lead to biased estimates and unreliable inference.

GPI departs from this paradigm by leveraging the internal
representations of GenAI models, which inherently encode rich semantic
information about the generated data and eliminate the need to
explicitly model or estimate representations. Further, GPI adopts a
fully nonparametric approach by using neural networks to estimate the
deconfounder, outcome model, and propensity score model, thereby
avoiding restrictive functional form assumptions. This combination of
principled representation learning and model flexibility enables more
robust causal inference in complex, high-dimensional settings.

Below, we demonstrate the wide applicability of GPI by reanalyzing three
empirical studies under the proposed framework: estimating the causal
effect of government censorship experience on the subsequent
censorship or self-censorship \citep{roberts_adjusting_2020},
estimating the effects of nighttime scenes on the perceived violence, and
estimating the persuasiveness of the political rhetoric
\citep{blumenau_variable_2022}.

\section{Text as Confounder}\label{sec:confounder}

Social scientists have documented that the Chinese government
selectively censors its citizens' social media posts
\citep[e.g.,][]{bamman2012censorship, king_how_2013}. Building on this
literature, \citet{roberts_adjusting_2020} investigated whether users
whose posts are censored are more likely to face future censorship and
whether they respond by engaging in self-censorship.

The authors analyzed data from 75,324 Weibo posts collected by the
Weiboscope project \citep{fu2013assessing}, which captures posts in
real time and later checks whether they have been removed.  Each post
serves as a focal post, and they constructed three outcome variables:
(i) the number of posts made by the same user in the four weeks
following the focal post, (ii) the proportion of those posts that were
censored during the same time period; and (iii) the proportion of
posts that went missing during the same four-week window.

Because censored and uncensored posts differ systematically in their
contents, a naive regression of these outcomes on the treatment
variable (prior censorship) is likely to yield biased causal
estimates. To address this potential confounding bias, the original
analysis adjusts for the content of the focal post based on a topic
model, the posting date, and each user's prior censorship history. We
adopt the same covariate adjustment strategy within the GPI framework.

We use two open-source large language models, LLaMA3 with eight
billion parameters \citep{grattafiori2024llama} and Gemma3 with one
billion parameters \citep{gemmateam2024gemmaopenmodelsbased}, to
reproduce each focal post and extract its internal
representation. These representations are then used to jointly
estimate the deconfounder and outcome model within a single neural
network architecture. A separate neural network is used to estimate
the propensity score, conditional on the estimated deconfounder and
other observed confounders. The deconfounder and outcome model are
fine-tuned using Optuna, an automated hyperparameter optimization
framework \citep{akiba_optuna_2019}.

To estimate the average treatment effect on the treated (ATT), we
apply double machine learning (DML) with two-fold cross-fitting
\citep{chernozhukov_doubledebiased_2018}. Standard errors are
clustered at the user level, and extreme propensity scores are
truncated following the method proposed by \citet{dorn2025much}.  We
also construct the optimal combination of the GPI estimates from the
two GenAI models based on their estimated influence functions, which
can yield more efficient estimates.  Theoretical justifications
and further implementation details are provided in Appendix
Sections~\ref{subsec:textasconfounder_theory}
and~\ref{subsec:textasconfounder_details}.

\begin{figure}[t]
  \centering \spacingset{1}
  \includegraphics[width=1.0\linewidth]{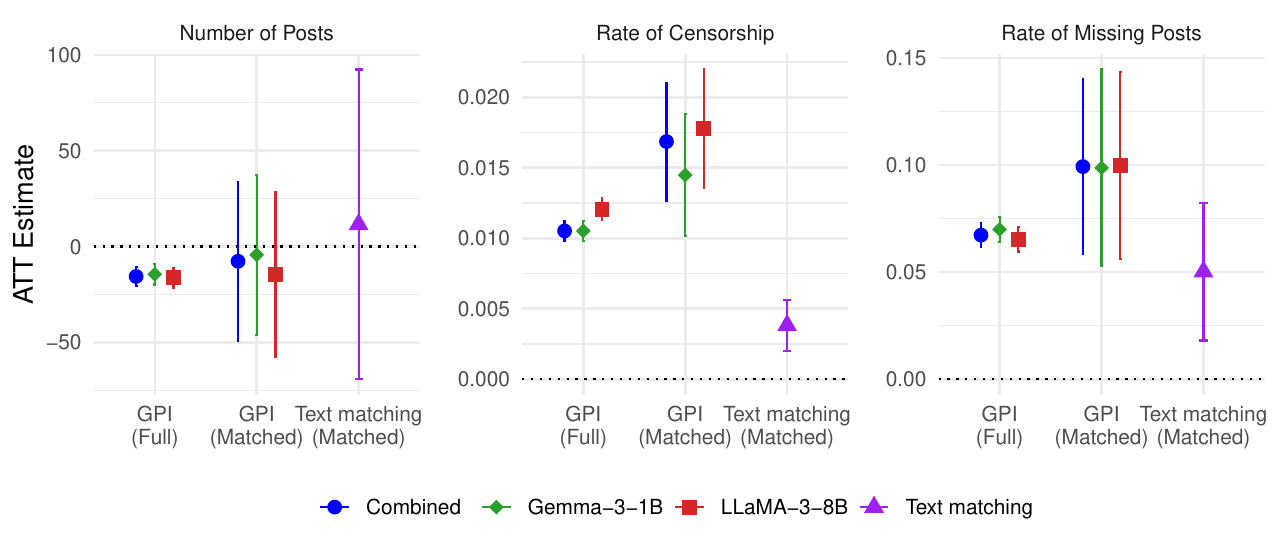}
  \caption{Estimated causal effects of prior censorship
    experience. For each outcome, we report the GPI estimates based on
    LLaMA3 with eight billion parameters (green diamonds) and Gemma3 with
    one billion parameters (red squares) using the full sample (Full)
    and the matched sample (Matched).  We also present the estimates
    from an optimally weighted combination of the two models (blue
    circles).  We compare GPI estimates with those of the text
    matching approach used in the original analysis (purple
    triangles).  The interval bars represent 95\% confidence intervals
    with the standard errors clustered at the user level. }
    \label{confounder_result}
\end{figure}

Figure~\ref{confounder_result} compares the GPI estimates with those
based on the text matching approach used in the original analysis,
which employed coarsened exact matching \citep{iacu:king:porr:11a} on
estimates from the structural topic model
\citep{roberts_model_2016}. For comparison, we apply GPI to both the
matched sample (628 users, 879 posts) and the full sample (4,155
users; 75,324 posts).

In contrast to the original analysis, GPI finds that prior censorship
significantly reduces users' subsequent posting activity, providing
evidence of self-censorship. Moreover, GPI reveals a much greater
effect of prior censorship on the likelihood of future censorship. In
addition, GPI's full-sample estimates are substantially more efficient
than those based on the matched sample, highlighting GPI's advantages
in statistical efficiency. These findings are consistent across both
LLaMA3 and Gemma3 models.  Lastly, optimally combining these two
results leads to a modest reduction in the standard error by 4.4\% on
average.

To further illustrate GPI's advantage, we assess robustness to a
text-based confounder by estimating each method with and without this
confounder and calculating the relative absolute bias, which is
defined as the absolute change in a point estimate relative to the
original estimate without the text-based confounders. The text-based
confounder of interest is measured as the proportion of 60 keywords
related to censorship or self-censorship as identified by
\citet{fu2013assessing}.  If a method effectively adjusts for the
confounding features in focal posts, the additional adjustment for the
text-based confounder should yield only a small change.  We use the
delta method for the calculation of GPI's standard errors.  For text
matching, we use bootstrap since an analytical standard error formula
is unavailable.

\begin{table}[!t]
\centering
\caption{Relative absolute bias due to a text-based confounder.
  Standard errors are in parentheses.  ``Full'' refers to estimates
  using the entire sample, while ``Matched'' refers to estimates based
  on the matched sample.  } \label{correlation}
\begin{tabular}{lccc}
    \toprule
    & \multicolumn{2}{c}{\textbf{GPI {\footnotesize (LLaMA3-8B)}}} &
                                                     \multicolumn{1}{c}{\textbf{Text matching}} \\
    \textbf{Outcome} & \multicolumn{1}{c}{Full} & \multicolumn{1}{c}{Matched} & \multicolumn{1}{c}{Matched} \\
    \midrule
Number of posts & 0.084 & 0.242 &  5.350 \\
                    & (0.068) & (0.682) & (33.794) \\
    Rate of censorship & 0.007 & 0.196 & 0.202 \\
                       & (0.003) & (0.140)  & (0.277) \\
    Rate of missing posts & 0.040 & 0.279 & 0.627 \\
                          & (0.049) & (0.346) & (3.416) \\
    \bottomrule
  \end{tabular}
\end{table}

As shown in Table~\ref{correlation}, GPI consistently produces smaller
relative absolute bias than text matching.  The results based on
Gemma3 are shown in Table~\ref{tab:alternative_results} in
Appendix~\ref{subsec:additional_results_text}.  We find that the GPI
estimates produce a similar ATT estimate for each outcome, regardless
of which LLM model is used.  We find that the optimal combination does
not substantially reduce the standard error in our application because
the estimated influence functions from LLaMA3-8B and Gemma3-1B are
highly correlated in our application as shown in Table~\ref{if_corr}
in Appendix~\ref{subsec:additional_results_text}. This high
correlation may be due to the fact that different LLMs are trained on
similar sets of data.

We investigate whether the advantage of GPI arises from text
representation, improved estimation, or both.  To do this, we
implement the same estimation procedure using two off-the-shelf text
embeddings, including Sentence-BERT \citep{reimers2019sentence} and
Qwen3 embedding with 0.6 billion parameters \citep{qwen3embedding}.
Unlike the internal representation of LLM, these alternative text
representations may not contain all relevant information necessary for
confounding adjustment.

The estimates based on these alternative representations are shown in
Table~\ref{tab:alternative_results} of
Appendix~\ref{subsec:additional_results_text}.  We find that the
estimated ATTs based on Qwen3-0.6B embeddings are qualitatively
similar to those of GPI, though the estimated ATE on the rate of
missingness is statistically significantly smaller than those of GPI.
In contrast, the results based on Sentence-BERT embeddings are quite
different, indicating that censorship decreases the rate of
subsequent censorship rather than increases it.  The estimated ATE on
the rate of missingness is even smaller.

We also conducted the above sensitivity analysis using these
alternative text representations.
Tables~\ref{correlation_app_gpi}~and~\ref{correlation_app_alternatives}
in Appendix~\ref{subsec:robustness} present the results.  We find that
the robustness of Qwen~3--0.6B embeddings is comparable to that of
GPI.  However, Sentence-BERT is much more sensitive to the inclusion
of additional textual confounders.  The corresponding estimates vary
much more as reflected by larger standard errors for the point
estimates of bias, some of which are also greater than those of GPI
and Qwen embeddings.  These results suggest that the quality of text
representation matters.  The advantage of GPI is that the internal
representation of LLM is guaranteed to contain {\it all} information
required for reproducing unstructured objects of interest.  This
theoretical guarantee comes at the computational cost as
pre-trained embedding models are typically less computationally
expensive than reproducing text through LLMs.

\section{Image as Treatment}\label{sec:image}

As protest images circulate widely on social media, they have become
an important source of information about the scale and character of
collective action. To study these images, \citet{won2017protest}
compiled a dataset of more than 9,000 protest images and obtained
human annotations for several attributes, including whether each image
depicts a daytime or nighttime scene and its perceived level of
violence.

We use this dataset to empirically validate the GPI methodology by
estimating the causal effect of nighttime scenes on perceived
violence. Nighttime protest images are associated with higher
perceived violence, as they are more likely to include visual cues
such as confrontations, fire, smoke, and police presence compared to
daytime images (correlation is 0.213).

However, because an image can be transformed from nighttime to daytime
without altering its underlying violent content, the causal effect of
nighttime on perceived violence is expected to be small; the
causal effect may not be zero given that the outcome variable is based
on human perception. This suggests that the GPI methodology should
substantially attenuate the association between nighttime and
perceived violence by adjusting for confounding features.

To apply the GPI methodology, we first reproduce each protest photo
using two different versions of Stable Diffusion model
(versions~1.5~and~2.1) \citep{rombach_high-resolution_2022} and
extract their internal representations. We then estimate the
deconfounder and outcome model using the same neural network
architecture and fine-tuning procedure as the one employed in the
previous application. Since the internal representation of images is
still relatively high-dimensional but still retains spatial structure, we use the
convolutional neural network (CNN) as an encoding layer.  Lastly, we
combine these estimates using the optimal weighting scheme.
Appendix~\ref{sec:ap_image} provides a theoretical justification of
GPI methodology for this application and the implementation details.

We compare the GPI methodology with (i) difference-in-means estimates
and (ii) estimates obtained from the same neural network architecture
applied directly to the original images, rather than to the internal
representations of Stable Diffusion models.

\begin{figure}[t]
  \centering \spacingset{1}
  \includegraphics[width=1.0\linewidth]{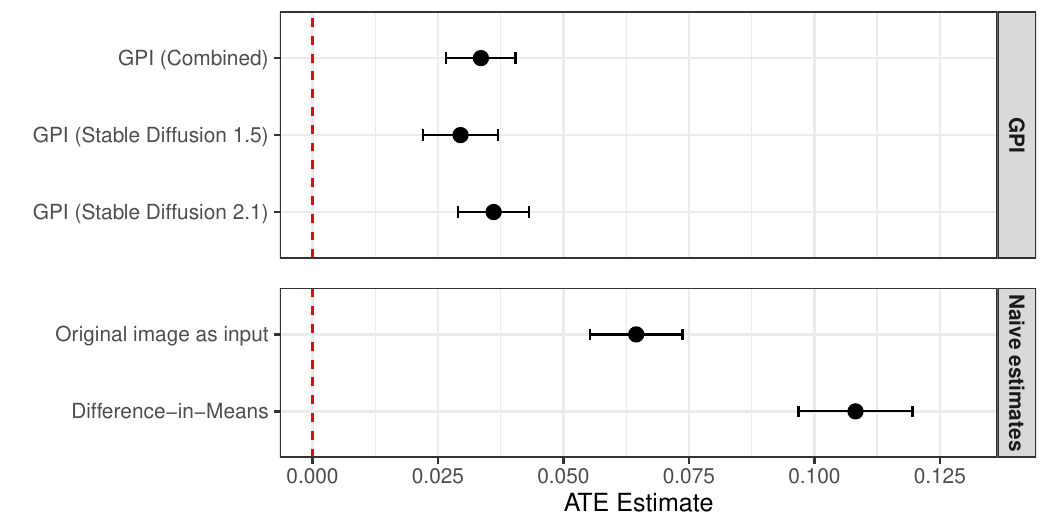}
  \caption{Estimated effects of nighttime scene on the perceived
    violence using GPI (top panel), directly using the original images
    with the same neural network architecture (first estimate in the
    bottom panel), and difference-in-means (second estimate in the
    bottom panel). Each bar represents the 95\% confidence
    interval.}
    \label{image_result}
\end{figure}

Figure~\ref{image_result} reports the results. Each horizontal bar
represents 95\% confidence interval. We find that the estimated effect
of nighttime scenes on perceived violence is reduced by more than half
once we adjust for other image features through the GPI methodology
(top panel). In addition, the estimates are similar across the two
different versions of Stable Diffusion model.  Combining the two
estimates reduces standard error by a modest amount of 7.4\% from GPI
(Stable Diffusion 1.5) and 2.2\% from GPI (Stable Diffusion 2.1).

Finally, the internal representation of Stable Diffusion model is of
substantially lower dimension ($16,384 = 64 \times 64 \times 4$) than
the original representation ($786,432 = 512 \times 512 \times 3$),
where all of our images have been adjusted to have an identical size
of $512 \times 512$ pixels.  The internal representation also contains
rich semantic information while the original representation only
includes positional and color information.  The bottom panel of
Figure~\ref{image_result} shows that the estimates based on the
original representation yield an estimate of much greater magnitude
than GPI, suggesting that the high-dimensionality and rich semantic
information of GenAI model's internal representation improve
empirical performance of causal effect estimation.  The GPI estimates
are also more precise, with substantially smaller standard errors than
the estimator based on the original representation.

\section{Structural Model of Text}\label{sec:text}

Political scientists and communication scholars have extensively
studied the effectiveness of different rhetorical strategies
\citep[e.g.,][]{jerit2009predictive,bos2013experimental}.  
Contributing to this literature, \citet{blumenau_variable_2022}
conducted a forced-choice conjoint experiment to evaluate the
persuasiveness of different types of political rhetoric. The authors
constructed 336 political arguments by systematically varying 12
policy issues, 14 rhetorical elements, and position (support or
opposition). Appendix~\ref{subsec:design} provides the full list of
policy issues and rhetorical features.

A total of 3,317 participants were each shown four randomly assigned
pairs of arguments. Within each pair, the arguments addressed the same
policy issue but took opposite sides and featured different, randomly
selected rhetorical elements. Participants were asked to indicate
which argument they found more persuasive or whether both were equally
persuasive, yielding a total of 13,268 pairwise comparisons. Appendix
Table~\ref{actual_arg} presents an example pair of arguments.

The authors estimated the latent persuasiveness of rhetorical elements
using a parametric structural model similar to the Bradley-Terry model
\citep{bradley1952rank} (see Appendix~\ref{subsec:model_persuasion}
for the details of the model). This original model assumes that the
probability of one argument being judged more persuasive than the other
is determined by an additive combination of three effects: an
interaction effect between policy area and position, the effect of the
rhetorical element, and a random effect for each individual
argument. The random effect is used to account for unobserved features
of the arguments that may influence persuasiveness but are not
explicitly measured. In addition, the original analysis manually
identified several potential confounders such as argument length and
included them as covariates.

We use GPI to enhance the original analysis by adjusting for the
confounding factors that are based on other features of arguments. We
compare the results based on three different LLMs: LLaMA3 with eight
billion parameters, along with two more recently released models ---
LLaMA3.3 with 70 billion parameters, and Gemma3 with one billion
parameters.  While all these models are instruction-tuned and thus can
be used to regenerate each argument and extract the internal
representation, their internal representations have varying dimensions
and contain different information.  Nevertheless, since these internal
representations are all sufficient for reproducing the original texts,
our theoretical results imply that the results based on these models
should be similar.

Once we extract internal representations, we then estimate a
{\it semiparametric} version of the original structural model,
which allows for greater flexibility in capturing complex relationships,
\begin{align*}
&\log \left[ \frac{\P(Y_{ijj'}  \leq k )}{\P(Y_{ijj'}  > k)} \right] = \delta_k + \beta_{T_j} - \beta_{T_{j'}} +  h(\bR_{j}) - h(\bR_{j'})
\end{align*}
where $Y_{ijj'} \in \{0,1,2\}$ denotes the relative persuasiveness of
arguments $j$ and $j'$ shown to respondent $i$ with $0$ indicating
that argument $j$ is more persuasive than $j'$, $1$ indicating equal
persuasiveness, and $2$ indicating argument $j$ is less persuasive.
The variable $T_j \in \{1,2,\ldots,14\}$ denotes a rhetorical element
used in argument $j$, and $h(\bR_j)$ represents latent confounding
features of that argument. The original analysis uses a parametric
version of this model, which may not capture all confounding features
of text.  We do not consider a fully nonparametric model given that
the number of unique text is only 336, making it impossible to
reliably estimate the causal effect of interest.

We estimate the latent persuasiveness parameter $\beta_{T_j}$ using a
neural network architecture for $h(\bR_j)$ and fine-tuning procedure
similar to those employed in the previous applications. This approach
allows us to evaluate the effect of each rhetorical element while
adjusting for latent confounding features embedded in the
arguments. To quantify uncertainty, we use bootstrap
with 200 replications. Theoretical details and
implementation specifics are provided in Appendix
Sections~\ref{subsec:model_persuasion}~and~\ref{subsec:implementation_persuasion},
respectively.

\begin{figure}[t]
  \centering \spacingset{1}
  \includegraphics[width=1.0\linewidth]{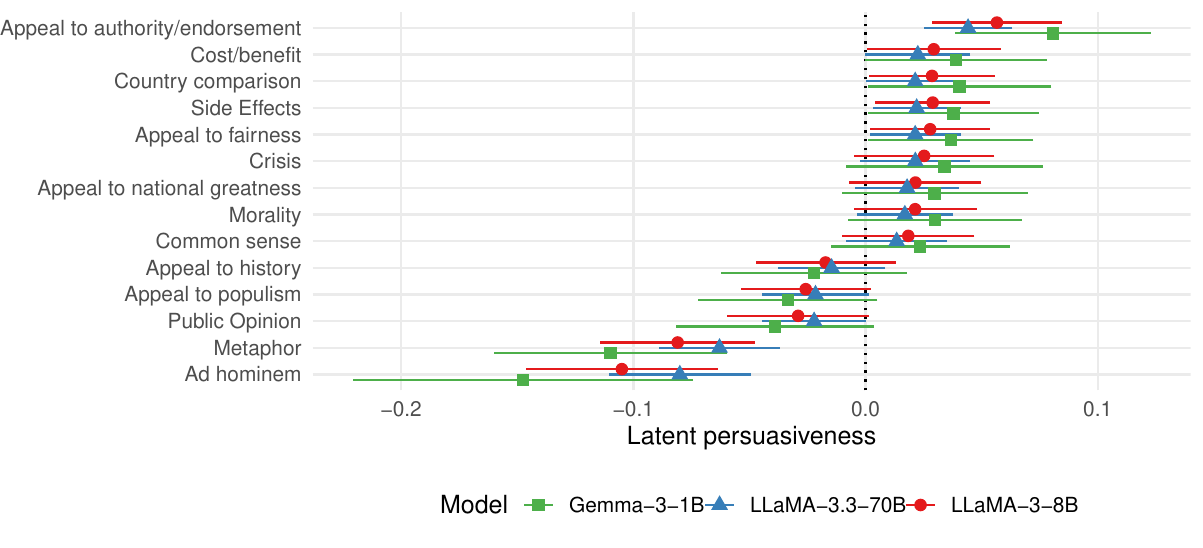}
  \caption{Estimated persuasiveness of 14 rhetorical elements in
    political arguments.  We present the estimated latent
    persuasiveness based on LLaMA3 with 8 billion parameters (red
    circle), LLaMA3.3 with 70 billion parameters (blue triangle),
    and Gemma3 with 1 billion parameters (green rectangle).  Each error
    bar represents the 95\% confidence intervals obtained by bootstrap.  }
    \label{texttreatment_result}
\end{figure}

Figure~\ref{texttreatment_result} presents the estimated effects of
the 14 rhetorical elements. We find that appeal to authority is the
most persuasive.  In contrast, ad hominem attacks, which represent
arguments targeting the person rather than their position,
significantly reduce persuasiveness. These findings are consistent
across GenAI models and each model's point estimates are highly
correlated (see Appendix Figure~\ref{fig:Text3_LLMpooledmean}).
While these findings are largely consistent with those of the original
analysis, the GPI estimates are substantially more precise. For
instance, the original results did not find statistically significant
effects for appeal to authority or cost/benefit arguments, despite
similar point estimates.

\section{Concluding Remarks}
\label{sec:conclusion}

In this paper, we introduced GenAI-Powered Inference (GPI), a
statistical framework that harnesses open-source GenAI models to
improve causal inference with unstructured data, such as text and
images. GPI extracts internal representations from these models,
yielding low-dimensional features that preserve all the relevant
information that is sufficient for generating the data.  By applying
machine learning to these representations, GPI enables robust
estimation of causal effects, while also quantifying estimation
uncertainty. We demonstrated GPI's broad applicability through three
different applications.

This work opens several promising avenues for future research. First,
developing methods to interpret the estimated deconfounder would help
researchers better understand the latent confounding features
identified by GPI. Second, GPI could be extended to discover effective
treatment features from unstructured data. While the current framework
assumes treatment features are predefined and observed, many
applications would benefit from discovering novel, more informative
treatments. Finally, although we focused on text and image data, the
GPI framework can be naturally extended to other types of unstructured
data, including audio and video \citep{nakamura2026causal,nakamura2026genai}.

\bigskip
\addcontentsline{toc}{section}{\refname}
\bibliography{references,my}

\newpage
\appendix

\setcounter{equation}{0}
\setcounter{figure}{0}
\setcounter{table}{0}
\setcounter{section}{0}
\renewcommand {\theequation} {S\arabic{equation}}
\renewcommand {\thefigure} {S\arabic{figure}}
\renewcommand {\thetable} {S\arabic{table}}
\renewcommand {\thesection} {S\arabic{section}}

\begin{center}
  {\LARGE \bf Supplementary Materials}
\end{center}

\section{A Simulation Study}\label{sec:simulation_studies}

We illustrate the intuition and performance of the proposed estimation
procedure through a simple simulation study with 200 Monte Carlo
replications.

\subsection{Simulation Setup}

We generate the data so that the treatment $T$ has highly restricted
support conditional on the representations $\bR$ but only part of the
representation $\bR$ acts as a confounder, which corresponds to
confounding feature $\bU$. As a result, the overlap assumption is
satisfied conditional on the confounding feature. Such settings arise
naturally with high-dimensional unstructured data such as text or
images, where features are numerous but many are irrelevant to the
outcome.

Specifically, for each observation $i \in \{1, \cdots, 5000\}$, we
generate the $d$-dimensional representations by
$\bR_{i} \sim \mathrm{MVN}(0, \bm \Sigma_d)$, where
$(\bm\Sigma_d)_{kk} = 1$ and $(\bm\Sigma_d)_{kj} = 0.1$ for
$j \neq k$, with $d = 4096$. We then split $\bR_i$ as
$$\bR_i = \begin{bmatrix} \bR_{i}^{(t)} \\ \bR_{i}^{(u)}
\end{bmatrix}$$
where
$\mathrm{dim}(\bR_{i}^{(t)}) = 16$ and $\mathrm{dim}(\bR_{i}^{(u)}) = 4080$.  We
then generate treatment $T_i$, confounder $\bU$, and outcome $Y_i$ as
\begin{align*}
    T_i = \mathbf{1}\{ \bm{1}_{16}^\top \bR_i^{(t)} > 0\}, \qquad \bU_i = \tanh(\bm{A}\bR_{i}^{(u)}) \qquad
    Y_i = \tau T_i + 20 \times  \bm{1}_{d_u} ^\top \bm U_{i} + \epsilon_i, \qquad \epsilon_i \sim \mathcal{N}(0, 1)
\end{align*}
where $d_u = \mathrm{dim}(\bU_i) = 16$, $\tau = 1$, and $\bm{A} \in \R^{16
  \times 4080}$ is a loading matrix where each entry is drawn from $\mathcal{N}(0,\frac{1}{4})$.

In this data generating process (DGP), the direct adjustment of
$\bR_i$ leads to the overlap violation, while conditioning on the
confounder does not. Our goal is to properly adjust
for confounder even though we do not directly observe it and must
estimate it from $\bR_i$. We generate the data 200 times, apply the
estimation procedure described below, and then calculate performance
metrics, including bias, root mean squared error (RMSE), and coverage
of the 95\% confidence interval.

\subsection{Estimation Procedure}

We compare four estimators on the simulated data: our GPI estimator
under two architectures and two baselines. Since we know the DGP, the
two GPI variants let us probe the role of architecture: (i) one sized
to capture exactly the necessary and sufficient part of $\bR$, and
(ii) one over-parameterized to capture most of $\bR$. The baseline
estimators are a naive application of augmented inverse probability
weighting with double machine learning (AIPW--DML)
\citep{chernozhukov_doubledebiased_2018} and DragonNet
\citep{shi_adapting_2019, veitch_adapting_2020}, which represent two
standard approaches for handling high-dimensional confounders but
neither of them addresses violation of the overlap assumption.

Specifically, we first apply the GPI estimation procedure, which is
designed to identify the treatment effect in this setting without
directly observing $\bU_i$. In short, GPI identifies the treatment
effect by adjusting for a deconfounder $\boldf(\bR_i)$ that satisfies
the mean independence condition
\begin{align}
    \E[Y_i \mid \bR_i, T_i] = \E[Y_i \mid \boldf(\bR_i), T_i]. \label{mean_independence}
\end{align}
We have shown that any function $\boldf(\bR_i)$ satisfying this mean
independence condition enables identification of the treatment effect.
See \citet{imai2024causal} and the application sections for the
theoretical properties of GPI.

To estimate the treatment effect, we use the neural network-based
estimation procedure proposed in \citet{imai2024causal}. Specifically,
we use the Treatment-Agnostic Representation Network (TarNet) of
\citet{shalit_estimating_2017}, which estimates the conditional
potential outcome function given the deconfounder:
\begin{align*}
  \mu_t(\boldf(\bR_i)) := \E[Y_i(t) \mid \boldf(\bR_i)]
  \quad \text{for } t = 0,1.
\end{align*}
Our architecture simultaneously estimates the deconfounder and the
outcome model by minimizing
\begin{align}
    \{ \hat{\bm{\lambda}}, \hat{\bm{\theta}}_0, \hat{\bm{\theta}}_1\}
    := \argmin_{{\bm{\lambda}}, {\bm{\theta}}_0, {\bm{\theta}}_1}
    \frac{1}{N} \sum_{i = 1}^N
    \bigl\{
    Y_i - \mu_{T_i}(\boldf(\bR_i; \bm\lambda);
    \bm\theta_{T_i})
    \bigr\}^2,
    \label{loss_func}
\end{align}
where $N$ denotes the sample size, and we make the neural network
parameters explicit: $\bm\lambda$ denotes the parameters of the
deconfounder $\boldf$, and $\bm\theta_t$ denotes the parameters of the
outcome regression function $\mu_t$. We then estimate the propensity
score model given the estimated deconfounder,
\[
\pi(\boldf(\bR_i; \hat{\bm\lambda}))
=
\Pr(T_i = 1 \mid \boldf(\bR_i; \hat{\bm\lambda})),
\]
and apply the double machine learning procedure using the estimated
outcome models $\{\hat\mu_1,\hat\mu_0\}$ and the estimated propensity
score model $\hat\pi$.

Using a deconfounder allows us to discard information in $\bR_i$ that
is predictive only of treatment assignment, thereby making the
positivity assumption more plausible. This feature also highlights the
importance of architectural choices. The deconfounder network must be
sufficiently expressive to retain the outcome-relevant variation in
$\bR_i$ and satisfy the mean independence condition in
\eqref{mean_independence}, while remaining sufficiently parsimonious
to avoid retaining variation that only predicts treatment
assignment. If the architecture is too restrictive, the outcome model
may fail to converge to the true population outcome function. If it is
too flexible or over-parameterized, however, the learned deconfounder
may preserve treatment-specific information from $\bR_i$, thereby
reintroducing the same overlap problem that arises when directly
adjusting for the full representation. Therefore, it is crucial to
select architectures and hyperparameters that achieve strong
out-of-sample outcome prediction performance, since treatment-specific
information that is irrelevant to the outcome should not improve
out-of-sample prediction.

We demonstrate the importance of choosing an appropriate architecture
by comparing an exact TarNet architecture with an over-parameterized
TarNet architecture. For the exact architecture, we use a one-layer
neural network with output dimension 1024 for the deconfounder and a
one-layer neural network with output dimension 1 for each outcome
head, using ReLU activation between layers. This architecture is not
over-parameterized because we know that the relationship between the confounding feature
and the outcome is linear.  We compare this exact architecture with an
over-parameterized architecture, which uses two consecutive layers
with dimensions 4096 for the deconfounder and two consecutive
layers with dimensions 10 and 1 for each outcome head.

For both architectures, we optimize
the network using Adam optimizer with learning rate 0.00002 and batch
size 256, and early stopping is
triggered if validation loss fails to improve for 5 consecutive
epochs. For propensity score estimation, we used a neural network with
spectral normalization \citep{gouk2021regularisation} to enforce
Lipschitz continuity. This network consisted of a two-layer MLP with
ReLU activation, using 256 hidden units in the first layer and 128 in
the second. The learning rate was fixed at $2 \times 10^{-5}$, and the model
was trained under the same early stopping criteria. We used a two-fold
cross-fitting procedure for the estimation.

We compare these GPI estimators with two baseline estimators. The first
is DragonNet \citep{shi_adapting_2019, veitch_adapting_2020},
which is a widely used neural network architecture when dealing with
unstructured data \citep{veitch_adapting_2020,
  gui_causal_2023}. DragonNet predicts the outcome using a shared
representation, while also encouraging this representation to be
predictive of treatment assignment. Specifically, it minimizes the
combined outcome and treatment-prediction loss function,
\begin{align}
    \frac{1}{N} \sum_{i = 1}^N
    \bigl\{
    Y_i - Q_{T_i}(\bm{b}(\bR_i))
    \bigr\}^2
    - \frac{1}{N} \sum_{i = 1}^N
    \biggl\{
    T_i \log g(\bm{b}(\bR_i))
    + (1 - T_i) \log[1 - g(\bm{b}(\bR_i))]
    \biggr\},
    \label{loss_func2}
\end{align}
where $Q_t(\cdot)$ is the outcome model under treatment status
$T_i=t$, $g(\cdot)$ is the treatment-prediction model, and
$\bm{b}(\bR_i)$ is the shared representation across treatment. To compare with the GPI
estimator, we use a one-layer neural network with output dimension 1024
for the shared representation, a one-layer network with output
dimension for each outcome head, and a two-layer network with dimension 256 and 128 for treatment head, and optimize the
neural network in the exact same way as the GPI estimator. We then use the
estimated outcome prediction $\{\hat{Q}_1, \hat{Q}_0\}$ and treatment
prediction $\hat g$ for the double machine learning (DML) procedure with
two-fold cross-fitting \citep{chernozhukov_doubledebiased_2018}.

Finally, we naively implement the DML procedure with the AIPW score
directly using the full representation $\bR_i$ as
covariates. Specifically, we estimate both the treatment and outcome
models using random forests with 100 trees, a minimum leaf size of 10,
and 50\% of the input features considered as candidate variables at
each split. We then estimate the treatment effect using two-fold
cross-fitting.

\subsection{Results}
\label{subsec:sim_results}

Table~\ref{tab:simulation_performance} reports the bias, RMSE,
and empirical coverage of the 95\% confidence interval. As expected, GPI with a properly
selected TarNet mitigates the positivity violation and eliminates the
resulting bias and achieves nominal coverage, whereas the two
baselines, DragonNet and AIPW--DML, suffer from severe bias. GPI with
an over-parameterized TarNet also exhibits non-negligible bias because
it fails to adequately reduce the dimensionality of the inputs. These
results highlight the importance of properly tuning the architecture
and hyperparameters of TarNet within the GPI framework.

\begin{table}[!t]
\centering
\caption{Simulation performance across 200 Monte Carlo
  replications. We report bias, root mean squared error (RMSE), and
  the empirical coverage of the 95\% confidence interval.}
\label{tab:simulation_performance}
\begin{tabular}{lrrr}
\toprule
Estimator & Mean Absolute Bias & RMSE & 95\% CI Coverage \\
\midrule
GPI (Exact) & 1.692  & 2.146  & 0.965 \\
GPI (over-parameterized) & 3.269  & 4.212 & 0.760 \\
DragonNet   & 3.440 & 4.291 & 0.850 \\
AIPW--DML         & 4.394 & 6.207 & 0.560 \\
\bottomrule
\end{tabular}
\end{table}

\section{Text as Confounder}\label{sec:appendix_textconfounder}

In this section, we provide a formal theoretical justification of our
GPI methodology introduced in Section~\ref{sec:confounder} of the main
text and its implementation details.

\subsection{Theoretical Properties}
\label{subsec:textasconfounder_theory}

Our approach extends the method proposed by \citet{imai2024causal} to
the case where text serves as confounder. We begin by defining the
causal quantity of interest.  We then establish its nonparametric
identification and develop an estimation strategy.

\subsubsection{Setup and Assumptions}
\label{subsec:setup_text}

Suppose we observe a sample of $N$ independent and identically
distributed (i.i.d.) units $i = 1, \ldots, N$ from a population of
interest. For each unit $i$, we observe a treatment assignment
$T_i \in \cT \subset \mathbb{R}$ and an observed outcome
$Y_i \in \cY \subset \mathbb{R}$, where $\cT$ and $\cY$ denote the
support of the treatment and that of the outcome,
respectively. Additionally, we observe a set of structured confounders
$\bm{Z}_i \in \mathcal{Z} \subset \mathbb{R}^d$ and unstructured
high-dimensional objects (e.g., texts or images)
$\bX_i \in \mathcal{X} \subset \mathbb{R}^r$, some features of which
also serve as confounders.

We adopt the potential outcomes framework for causal inference and
assume the standard consistency assumption
\citep{rubi:90}. Specifically, we assume that the potential outcome,
denoted by $Y_i(t)$, depends solely on the treatment assignment of
unit $i$ itself, not on those of other units. The assumption is
formalized as follows:
\begin{assumption}[Consistency]\label{consistency_confounder} The potential outcome under the treatment $t \in \cT$ is denoted by $Y_i(t)$, and equals the observed outcome $Y_i$ under the realized treatment assignment $T_i$:
\begin{align*}
  Y_i = Y_i(T_i).
\end{align*}
\end{assumption}

We are interested in estimating the marginal mean of the potential
outcome under treatment condition $T_i = t$ for some $t \in \cT$
(i.e., $\E[Y_i(t)]$), which can be used to estimate the average
treatment effect (ATE). We consider the strong latent ignorability
assumption; conditional on the observed covariates and certain unknown
features of the high-dimensional unstructured objects, potential
outcomes are independent of treatment assignment. This assumption
guarantees the existence of such latent confounding features.
\begin{assumption} {\sc (Strong Latent Ignorability with Unknown
    Confounding Features)}\label{ignorability_confounder} There exists a
deterministic function $g_{\bU}: \cX \to \cU$ that maps an
unstructured object $\bX_i$ onto the low-dimensional confounding
features $\bU_i \in \cU$ with $\dim(\bU) \ll \dim(\bX)$ such that the
potential outcomes are independent of the treatment assignment given
the observed covariates $\bm{Z}_i$ and the confounding features
$\bU_i= g_{\bU}(\bX_i)$:
\begin{align*}
  \{Y_i(t)\}_{t \in \cT} \; \indep \; T_i \mid \bm{Z}_i = \bm{z},
  \bU_i = \bu,
\end{align*}
where $\P(T_i = t \mid \bm{Z}_i = \bm{z}, \bU_i = \bu) > 0$ for all
$t \in \cT$, $\bm{z} \in \mathcal{Z}$, and $\bu \in \cU$.
\end{assumption}

Importantly, we assume the existence of low-dimensional features
$\bU_i$, which are deterministic functions of the unstructured objects
and sufficient to satisfy the strong ignorability assumption. We do
not condition directly on the high-dimensional unstructured objects
$\bX_i$ because doing so can often lead to perfect prediction of
treatment assignment, thereby violating the positivity condition and
yielding severely biased estimates \citep{d2021overlap}.

Prior work
either overlooks this issue \citep[e.g.,][]{veitch_adapting_2020,
  klaassen_doublemldeep_2024} or addresses it by using text-based
matching \citep[e.g.,][]{roberts_adjusting_2020,mozer_matching_2020}.
Matching is not well suited for high-dimensional controls and the
reliance on parametric propensity score models such as topic models
leads to bias. In contrast, our method assumes only the existence of
low-dimensional confounding features $\bU_i$.

Finally, we assume that unstructured objects are generated by a deep
generative model. When the existing images and text are of interest,
we can reproduce them by appropriately prompting a deep generative
model. Following \citet{imai2024causal}, we adopt a broad definition of
deep generative model to encompass LLMs and other foundation models.
\begin{definition}[Deep Generative Model]\label{deep_use} A deep
  generative model is the following possibly degenerate probabilistic model that takes
  prompt $\bP_i$ as an input and generates the unstructured object
  $\bX_i$ as an output:
$$
\begin{aligned}
&\P(\bX_i  \mid \bh_{\bgamma}(\bR_i))\\
&\P(\bR_i \mid \bP_i)
\end{aligned}
$$
where $\bR_i \in \cR \subset \R^{d}$ denotes an internal
representation of $\bX_i$ contained in the model and
$\bh_{\bgamma}(\bR_i)$ is a deterministic function parameterized by
$\bgamma$ that completely characterizes the conditional distribution
of $\bX_i$ given
$\bR_i$.  
\end{definition}

Our definition encompasses a wide range of existing text and image
generation models. Under this definition of deep generative model,
$\bR$ is a lower-dimensional representation of the unstructured object
$\bX$ and is also a hidden representation of neural networks.  We
assume that the last layer of the deep generative model is a
deterministic function of the internal representation $\bR_i$ so that
the low-dimensional features of the unstructured object $\bU_i$ can be
regarded as a deterministic function of the low-dimensional internal
representation $\bR_i$ of the deep generative model. The assumption
explicitly writes a generative model with a hierarchical structure to
emphasize that we can use any intermediate layer before the final
layer $\bm{h}_{\bm \gamma}(\bR_i)$ so long as $\bU_i$ is a
deterministic function of selected layer.

\begin{assumption}[Deterministic Decoding]\label{det_dec} \spacingset{1} The output layer of a deep generative model is deterministic.  That is, $\P(\bX_i  \mid \bh_{\bgamma}(\bR_i))$ in Definition~\ref{deep_use} is a degenerate distribution.
\end{assumption}

Crucially, we only require that $\bR_i$ deterministically generates
the unstructured object $\bX_i$. For example, some generative models,
especially diffusion models, have an internal architecture that is
stochastic, but the output of the decoder layer can be made
deterministic so that GPI is still applicable.  In the case of LLMs,
\citet{barrie2024replication} shows that due to updates to the internal
parameters, the outputs of LLMs are not generally replicable over
time. However, in this context, we are only concerned with the
deterministic relationship between $\bR_i$ and $\bX_i$, which can be
controlled by selecting appropriate hyperparameters for deterministic
decoding under any given LLM.

We emphasize that researchers can apply GPI to existing text or image
data by prompting deep generative models to reproduce the original
unstructured inputs as faithfully as possible. For text data, this can
be accomplished using instruction-tuned models with a system prompt
that enforces verbatim reproduction. For example, in our text
applications \citep{imai2024causal, nakamura2026genai}, we used the
prompt: ``You are a text generator who simply repeats the input
text,'' which consistently led LLMs to reproduce the original text
verbatim. Although exact reproduction may occasionally fail, such
cases can be readily detected by directly comparing the generated
output with the original text.

A similar strategy can be applied to images using diffusion
models. Because diffusion models are inherently stochastic, however,
the regenerated image is typically only approximately identical to the
original. In these settings, the resulting internal representations
should likewise be interpreted as approximations. Importantly, deep
generative models enable researchers to directly assess the quality of
these approximate representations by visually inspecting how
accurately the unstructured data are reconstructed.

\begin{figure}[t]
\centering \spacingset{1}
\includegraphics[width=0.7\textwidth]{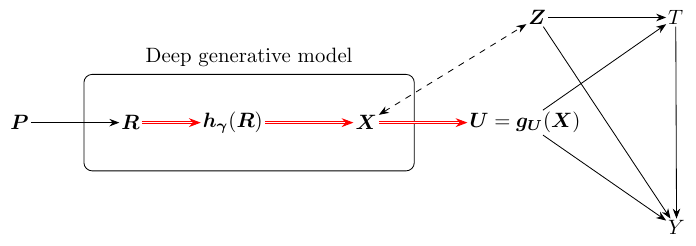}
\caption{Directed Acyclic Graph of the Assumed Data-Generating Process
  with Unstructured Confounder. An unstructured object $\bX$ (e.g., an
  image or text) is generated via a deep generative model (shown as a
  box), where a prompt $\bP$ produces an internal representation
  $\bR$, which is then transformed into $\bX$ by the function
  $\bh_{\bgamma}(\bR)$. The causal relationship between treatment $T$
  and outcome $Y$ is confounded by both observed structured covariates
  $\bm{Z}$ and latent confounding features $\bU$ embedded within the
  unstructured object $\bX$. Arrows with red double lines represent
  deterministic causal relationships while single-lined arrows denote
  possibly stochastic relationships and dashed bidirectional arrows
  indicate potential independence.}
\label{DAG1}
\end{figure}

Figure~\ref{DAG1} presents a directed acyclic graph (DAG) that
summarizes the data generating process and assumptions described
above. In this DAG, an arrow with double red-colored lines represents
a deterministic causal relation while an arrow with a single line
represents a stochastic causal relation. The dashed line bidirectional
arrow represents the potential independence relation.

\subsubsection{Identification}

Given this setup, we establish the nonparametric identification of the
mean of the potential outcome under the treatment
condition $T_i = t$ for any given $t \in \cT$. We extend the results
of \citet{imai2024causal} to the case of unstructured objects as
confounder, accommodating potentially non-binary discrete treatments and
observed structured confounders.

Specifically, we show that there exists a deconfounder function
$\boldf(\bR_i)$ satisfying the mean independence relation
$\E[Y_i \mid \bR_i, T_i, \bm{Z}_i, \boldf(\bR_i)] = \E[Y_i \mid
T_i, \bm{Z}_i, \boldf(\bR_i)]$, where $\boldf(\bR_i)$ is a
deterministic function of the internal representation $\bR_i$. The
existence of such a function is guaranteed by Assumptions
\ref{ignorability_confounder} and \ref{det_dec}, since $\bU_i$ is a
deterministic function of $\bR_i$ and satisfies the mean independence
condition. Moreover, any deconfounder function that satisfies the same
mean independence relation leads to the same identification formula.
The proof is given in Appendix~\ref{proof:prop_identification}.

\begin{proposition}[Nonparametric Identification]\label{prop:identification}
  Under
  Assumptions~\ref{consistency_confounder}--\ref{det_dec},
  there exists a deconfounder function
  $\boldf: \mathbb{R}^d \mapsto \mathbb{R}^q$ with $q \leq d$ that
  satisfies the following mean independence relation:
\begin{align}
\E[Y_i \mid \bR_i, T_i, \bm{Z}_i, \boldf(\bR_i)] = \E[Y_i \mid T_i, \bm{Z}_i, \boldf(\bR_i)]. \label{eq:deconfounder_meanindependence}
\end{align}
By adjusting for this deconfounder, we can nonparametrically identify
the mean potential outcome under the treatment condition $T_i = t$ as,
\begin{align}
  \E[Y_i(t)] = \int_{\mathcal{R} \times \mathcal{Z}}  \E\bigl[Y_i \mid T_i = t, \bm{Z}_i, \boldf(\bR_i)\bigr] dF(\bm{Z}_i, \bR_i). \label{eq:deconfounder}
\end{align}
\end{proposition}

\subsubsection{Estimation and Inference}
\label{subsec:estimation_text_confounder}

Given the identification formula presented in
Proposition~\ref{prop:identification}, we now turn to estimation and
statistical inference. We extend the estimation procedure of
\citet{imai2024causal} to the current setting. Our strategy relies on
two key observations. First, Assumption~\ref{ignorability_confounder}
implies that the deconfounder should not vary across the treatment
conditions. Second, the deconfounder must satisfy the conditional
independence stated in \eqref{eq:deconfounder_meanindependence}.

\begin{figure}[t]
\centering \spacingset{1}
\includegraphics[width=0.6\textwidth]{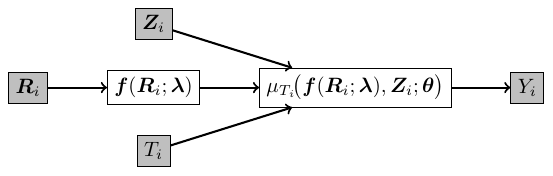}
\caption{Proposed Model Architecture with Unstructured Object as
  Confounder. The proposed model takes an internal representation of
  unstructured object $\bR_i$ as an input, and estimates a
  deconfounder $\boldf(\bR_i)$, which is a lower-dimensional
  representation of $\bR_i$, and then uses it together with the
  treatment $T_i$ and structured confounder $\bm Z_i$ to predict the
  conditional expectation of outcome
  $\mu_{T_i}(\boldf(\bR_i), \bm{Z}_i) := \E[Y_i \mid
  T_i,\boldf(\bR_i), \bm{Z}_i]$. The architecture directly encodes the
  mean independence relation
  $\E[Y_i \mid \bR_i, T_i, \bm{Z}_i, \boldf(\bR_i)] = \E[Y_i \mid
  T_i, \bm{Z}_i, \boldf(\bR_i)]$.}
\label{tarnet}
\end{figure}

We use a neural network architecture that extends TarNet \citep{shalit_estimating_2017}
to non-binary treatment and structured confounding
variables to estimate the conditional potential outcome function given
the deconfounder and observed control variables, i.e.,
\begin{align*}
  \mu_{t}(\boldf(\bR_i), \bm{Z}_i) = \E[Y_i(t) \mid \boldf(\bR_i), \bm{Z}_i].
\end{align*}
Our proposed neural network architecture, which is summarized in
Figure~\ref{tarnet}, simultaneously estimates the deconfounder
$\boldf(\bR_i)$ and the conditional potential outcome function
$\mu_{T_i}(\boldf(\bR_i), \bm{Z}_i):=\E[Y_i \mid T_i, \boldf(\bR_i),
\bm{Z}_i]$, by first projecting the internal representation $\bR_i$
onto a lower dimensional space of deconfounder $\boldf$.  We then use
the deconfounder, treatment $T_i$, and observed structured confounders
$\bm{Z}_i$ to predict the conditional potential outcome function
$\mu_{T_i}$ by minimizing the following loss function:
\begin{align}
  \{\hat\blambda, \hat\btheta\} = \argmin_{\blambda, \btheta} \frac{1}{N} \sum_{i=1}^N \{ Y_i - \mu_{T_i}(\boldf(\bR_i; \blambda), \bm{Z}_i; \btheta) \}^2, \label{loss_text_original}
\end{align}
where $\blambda$ is the parameter vector of deconfounder function
$\boldf(\bR_i)$, and $\btheta$ is the parameter vector of conditional
potential outcome function $\mu_{T_i}(\boldf(\bR_i), \bm{Z}_i)$. We
prevent the estimated deconfounder from varying across the treatment
levels by using the shared representation $\boldf(\bR_i)$ across treatment levels.

Importantly, GPI differs from the existing methods of causal inference
for texts such as \citet{veitch_adapting_2020} that directly predict
treatment assignment using text representation.  Doing so often leads
to the violation of the overlap assumption because each text is unique
and can perfectly predict treatment assignment.  Under this standard
approach, adding a prediction head for the treatment assignment
encourages the deconfounder to capture nearly all the information
predictive of treatment, thereby
exacerbating the violation of overlap assumption.
GPI avoids this problem by optimizing only the outcome model for the
best out-of-sample prediction, discarding the information that is
predictive only of the treatment. While the optimization problem in
\eqref{loss_text_original} does not explicitly penalize overlap
violations, we address them by learning a low-dimensional
representation (i.e., deconfounder), which is only predictive of the
outcome, through dimensionality reduction in the neural network
architecture and careful hyperparameter tuning (see
Appendix~\ref{sec:simulation_studies} for a simulation study that
illustrates this point).

Given the proposed neural network architecture, we estimate the
Average Treatment Effect for the Treated (ATT) using double machine
learning (DML) \citep{chernozhukov_doubledebiased_2018}. We estimate
ATT to compare our result with the original analysis based on text
matching.  Here, we propose to estimate the propensity score model as
a function of the estimated deconfounder, i.e.,
$\pi_t(\boldf(\bR_i), \bm{Z}_i) = \P(T_i = t \mid \boldf(\bR_i),
\bm{Z}_i)$ so that we do not directly condition on the
high-dimensional unstructured objects.

To ensure the asymptotic normality discussed later in this section,
the propensity score model must be Lipschitz-continuous; accordingly,
we employ a neural network with spectral normalization
\citep{gouk2021regularisation}. Practical implementation requires
careful tuning of the hyperparameters for the deconfounder and the outcome models,
including network width and depth, learning rate, batch size, dropout rate,
and number of epochs.  We automate this with advanced hyperparameter-optimization tools such as Optuna
\citep{akiba_optuna_2019}.

As an example, we present the entire estimation procedure in the case
of binary treatment $\cT = \{0,1\}$. Denote the observed data by
$\cD:=\{\cD_i\}_{i=1}^N$ where
$\cD_i := \{Y_i, T_i, \bR_i, \bm{Z}_i\}$. We use the following
$K$-fold cross-fitting procedure, assuming that $N$ is divisible by
$K$.
\begin{enumerate}
    \item Randomly partition the data into $K$ folds of equal size where the size of each fold is $n = N/K$.  The observation index is denoted by $I(i)\in \{1,\dots,K\}$ where $I(i)=k$ implies that the $i$th observation belongs to the $k$th fold.
    \item For each fold $k \in \{1, \cdots, K\}$, use observations with $I(i)\ne k$ as training data:
    \begin{enumerate}
    \item further split the training data into two folds, $I_1^{(-k)}$ and $I_2^{(-k)}$
    \item {\sloppy simultaneously obtain an estimated deconfounder and an estimated conditional outcome function on the first fold, which are denoted by $\hat\boldf^{(-k)}(\{\bR_i, \bm{Z}_i\}_{i \in I_1^{(-k)}}) :=$ $\boldf(\{\bR_i, \bm{Z}_i\}_{i \in I_1^{(-k)}}; \hat\blambda^{(-k)})$ and $\hat{\mu}_t^{(-k)}(\{\bR_i, \bm{Z}_i\}_{i \in I_1^{(-k)}}) :=$ $\mu_t(\boldf(\{\bR_i\}_{i \in I_1^{(-k)}}, \bm{Z}_i; \hat\blambda^{(-k)}); \hat\btheta^{(-k)})$, respectively, by solving the optimization problem given in Equation~\eqref{loss_text_original}, and\par}
    \item {\sloppy obtain an estimated propensity score given the estimated deconfounder on the second fold, which is denoted by $\hat{\pi}^{(-k)}(\hat\boldf^{(-k)}(\{\bR_i\}_{i \in I_2^{(-k)}}), \bm{Z}_i) :=$ $\hat{\pi}^{(-k)} (\boldf(\{\bR_i\}_{i \in I_2^{(-k)}}; \hat\blambda^{(-k)}), \bm{Z}_i)$.\par}
    \end{enumerate}
  \item Compute the ATT estimator $\hat{\tau}$ as a solution to:
    \begin{equation}
      \frac{1}{nK}\sum_{k = 1}^K  \sum_{i: I(i) =k} \psi(\cD_i; \hat{\tau}, \hat{\boldf}^{(-k)}, \mu^{(-k)}_0, \hat\pi^{(-k)}) \ = \ 0, \label{eq:tau.hat}
      \end{equation}
      where
      \begin{equation}
        \begin{aligned}
        & \psi(\cD_i; \tau, \boldf, \mu_0, \pi)\\
        & \ = \  \frac{T_i (Y_i - \mu_0(\boldf(\bR_i), \bm{Z}_i))}{\P(T_i = 1)} -
        \frac{\pi(\boldf(\bR_i), \bm{Z}_i) (1 - T_i) (Y_i - \mu_0(\boldf(\bR_i), \bm{Z}_i) ) }{\P(T_i = 1)(1 - \pi(\boldf(\bR_i), \bm{Z}_i))}\\
        & \qquad - \frac{T_i \tau}{\P(T_i = 1)}.
        \end{aligned} \label{eq:psi}
      \end{equation}
\end{enumerate}

Finally, we establish the asymptotic properties of the proposed
estimator in the case of binary treatment. We assume that the
following regularity conditions hold. These conditions are similar to
those in \citet{imai2024causal} except that we also condition on the
observed covariates $\bm{Z}_i$.
\begin{assumption}[Regularity Conditions]\label{reg_text} \spacingset{1} Let $c_1$, $c_2$, and $q > 2$ be positive constants and $\delta_n$ be a sequence of positive constants approaching zero as the sample size $n$ increases.  Then, the following conditions hold.
  \begin{enumerate}
    \item[(a)] \label{reg_text_primitive} (Primitive conditions)
    \begin{align*}
        \E[|Y_i|^q]^{1/q} \leq c_1, &\quad \sup_{\br \in \mathcal{R}, \bm{z} \in \mathcal{Z}}\E [|Y_i - \mu_{T_i}(\boldf(\br), \bm{z} )|^2 \mid \bR_i = \br, \bm{Z}_i = \bm{z}] \leq c_1,\\
        &\E[ |Y_i - \mu_{T_i}(\boldf(\bR_i), \bm{Z}_i)|^{2} ]^{1/2} \geq c_2.
    \end{align*}
  \item[(b)] (Outcome model estimation) \label{reg_text_outcome}
    \begin{align*}
        &\E[|\hat{\mu}_{T_i}(\hat{\boldf}(\bR_i), \bm{Z}_i) - \mu_{T_i}(\boldf(\bR_i), \bm{Z}_i)|^q]^{1/q} \leq c_1, \\
        &\E[|\hat{\mu}_{T_i}(\hat{\boldf}(\bR_i), \bm{Z}_i) - \mu_{T_i}(\boldf(\bR_i), \bm{Z}_i)|^2]^{1/2} \leq \delta_n n^{-1/4}.
    \end{align*}
  \item[(c)] (Deconfounder estimation) \label{reg_text_deconfounder}
   \begin{align*}
      & \E\left[\norm{\hat{\boldf}(\bR_i) - \boldf(\bR_i)}^q\right]^{1/q} \leq c_1, \quad \mathbb{E}\left[ \norm{\hat{\boldf}(\bR_i) - \boldf(\bR_i)}^2 \right]^{1/2} \leq \delta_n n^{-1/4}
    \end{align*}
  \item[(d)] (Propensity score estimation) \label{reg_text_propensity} $\pi(\cdot)$ is Lipschitz continuous at the every point of its support, and satisfies:
    \begin{align*}
        & \E[|\hat{\pi}(\boldf(\bR_i), \bm{Z}_i) - \pi(\boldf(\bR_i), \bm{Z}_i)|^q]^{1/q} \leq c_1, \\
        & \E[|\hat{\pi}(\boldf(\bR_i), \bm{Z}_i) - \pi(\boldf(\bR_i), \bm{Z}_i)|^2]^{1/2} \leq \delta_n n^{-1/4}.
    \end{align*}
  \end{enumerate}
\end{assumption}
These regularity conditions are standard in the DML literature, and it
is known that the neural network architecture with an appropriate
depth and width can achieve the required convergence rates
\citep{farrell_deep_2021}. The propensity score requires the Lipschitz
continuity condition, which can be satisfied by using a regularized
neural network architecture \citep{gouk2021regularisation}.

Given the above assumptions, the asymptotic normality of the proposed
estimator follows immediately from the DML theory.
\begin{theorem}[Asymptotic Normality of the ATT Estimator]\label{asymp_text} \spacingset{1}
Under Assumptions~\ref{consistency_confounder}--\ref{reg_text}, the estimator $\hat{\tau}$ obtained from the influence function $\psi$ satisfies asymptotic normality:
$$
\frac{\sqrt{N}(\hat{\tau} - \tau)}{\sigma} \xrightarrow[]{d} \mathcal{N}(0, 1)
$$
where $\sigma^2 = \E[\psi(\cD_i; \tau, \boldf, \mu_0, \pi)^2].$
\end{theorem}
We omit the proof as it is identical to that of Theorem~1 given in \citet{imai2024causal}.

When we have internal representations from different GenAI models, we can
combine the estimates based on them to improve statistical efficiency.
The following proposition establishes the asymptotic normality of such
combined estimator in the case of the two GenAI models.  The extension
to more than two GenAI models is straightforward.
\begin{proposition}[Asymptotic Normality of the Combined
  Estimator] \label{asymp_combined} Suppose that we have two different
  GenAI models that give two different internal representations. Let
  $\cD_{ij} = \{Y_i, T_i, \bR_{ij}, \bm{Z}_i\}$ be the observed data
  obtained from the $j$th GenAI model for $j=1,2$, where $\bR_{ij}$ is
  the internal representation obtained from the $j$th GenAI model. Let
  $\hat{\tau}_j$ be the ATT estimators obtained from the $j$th GenAI
  model and
  $\psi_j(\cD_{ij}) = \psi_j(\cD_{ij}; \tau, \boldf_{j}, \mu_{1j},
  \mu_{0j}, \pi_{1j})$ be the influence function for $\hat{\tau}_j$
  for $j=1,2$, where $\boldf_j$, $\mu_{tj}$, and $\pi_{1j}$ are the
  true deconfounder, outcome model, and propensity score model for the
  $j$th GenAI model, respectively. Under
  Assumptions~\ref{consistency_confounder}--\ref{reg_text} for each
  GenAI model, the combined estimator
  $\hat{\tau}_{\rm comb} = \omega \hat{\tau}_1 + (1-\omega)
  \hat{\tau}_2$ for any weight $\omega$ satisfies asymptotic
  normality:
$$
\sqrt{N}\frac{(\hat{\tau}_{\rm comb} - \tau)}{\sigma_{\rm comb}} \xrightarrow[]{d} \mathcal{N}(0, 1)
$$
where $\sigma_{\rm comb}^2 = \E[(\omega\psi_1(\cD_{i1}) + (1-\omega)\psi_2(\cD_{i2}))^2]$.
Moreover, the combined estimator achieves the minimum asymptotic variance at the optimal weight
$$
\omega^* = \frac{\E[\psi_2(\cD_{i2})^2] - \E[\psi_1(\cD_{i1})\psi_2(\cD_{i2})]}{\E[\psi_1(\cD_{i1})^2] + \E[\psi_2(\cD_{i2})^2] - 2\E[\psi_1(\cD_{i1})\psi_2(\cD_{i2})]}.
$$
\end{proposition}
The proof is given in Appendix~\ref{proof:asymp_combined}. The
combined estimator can substantially reduce variance when the
correlation between the two influence functions is low, indicating
that the different representations provide complementary information
about the treatment effect.

\subsection{Implementation Details for the Empirical Application}
\label{subsec:textasconfounder_details}

We use the dataset analyzed by \citet{roberts_adjusting_2020}, which is
based on the data from the Weiboscope project
\citep{fu2013assessing}. This project collected Weibo posts in real
time and subsequently revisited them to determine whether they had
been censored. The original analysis focused on users who experienced
at least one instance of censorship during the first half of 2012, and
examined their subsequent posts in the second half of that year. To
construct the control group, the authors selected posts that had a
cosine similarity greater than 0.5 to a censored post and were posted
on the same day. Posts shorter than 15 characters were excluded. The
final dataset we analyze consists of 75,324 posts from 4,155 Weibo
users.

Based on the treatment and control ``focal'' posts, three outcome
variables were constructed:
\begin{enumerate}
\item The number of posts made by the same user within the four weeks following the focal post
\item The proportion of those posts that were censored (as indicated
  by a ``permission denied'' message)
\item The proportion of posts that went missing (as indicated by a ``Weibo does not exist'' message)
\end{enumerate}
It is important to note that the second and third outcomes capture
different types of post removals. As noted by \citet{fu2013assessing},
a ``permission denied'' message is a clear signal of censorship,
whereas ``Weibo does not exist'' may reflect either censorship or
voluntary deletion by the user. To adjust for users' baseline
behavior, the same three measures were also computed for the four-week
period preceding each focal post and included as confounding variables.

We apply the GPI methodology. First, we use LLaMA3 (eight billion
parameters) and Gemma3 (one billion parameters) to reproduce all posts
in the dataset and extract the internal representation of the last
token. Due to the autoregressive nature of generative models, this
final-token representation captures most of the information of each
post \citep{neelakantan2022text, ma2024fine, jiang2024scaling}. The
resulting internal representation, denoted by $\bR_i$, has a
dimensionality of 4,096 for LLaMA3 and 1,152 for Gemma3. Using these
representations, we estimate the deconfounder $\boldf(\bR_i)$ and the
outcome model $\mu_{T_i}(\boldf(\bR_i), \bm{Z}_i)$ via a neural
network architecture described in the previous section.

To train the model for each outcome variable, we employed two-fold
cross-validation on the full sample, following the recommendation of
\citet{bach2024hyperparameter}. The hyperparameter search space
includes:
\begin{itemize}
\item Learning rate: $[10^{-7}, 10^{-4}]$
\item Dropout rate: $[0.05, 0.3]$
\item Outcome model architecture: single-layer MLP with ReLU activation and 50, 100, or 200 hidden units
\item Deconfounder architecture: two-layer MLP with ReLU activation and hidden unit configurations of (256, 128), (512, 256), or (1024, 512)
\end{itemize}

\begin{table}[t]
\centering
\caption{Optimal hyperparameters selected for each outcome measure using Optuna. The table lists the learning rate, dropout rate, outcome model architecture, and deconfounder architecture applied to the reanalysis of \citet{roberts_adjusting_2020}.}
\label{tab:hyperparameters}
\spacingset{1}
\resizebox{\textwidth}{!}{%
\begin{tabular}{llcccc}
\toprule
\textbf{Model} &\textbf{Outcome} & \textbf{Learning Rate} & \textbf{Dropout} & \textbf{Outcome Model} & \textbf{Deconfounder} \\
\midrule
LLaMA3& Number of Posts & $1.188 \times 10^{-7}$ & 0.051 & [100, 1] & [1024, 512] \\
(8B) & Rate of Censorship & $9.974 \times 10^{-5}$ & 0.161 & [100, 1] & [1024, 512] \\
& Rate of Missing Posts & $9.858 \times 10^{-5}$ & 0.118 & [50, 1] & [1024, 512] \\
\midrule
Gemma3 & Number of Posts & $2.726 \times 10^{-7}$  & 0.094 & [100, 1] & [1024, 512] \\
(1B) & Rate of Censorship & $7.678 \times 10^{-5}$ & 0.276 & [50,1] & [1024, 512] \\
& Rate of Missing Posts & $7.499 \times 10^{-5}$ & 0.154 & [50,1] & [512, 256] \\
\midrule
Qwen & Number of Posts & $6.100 \times 10^{-5}$ & 0.160 & [100, 1] & [1024, 512] \\
(0.6B) & Rate of Censorship & $9.679 \times 10^{-5}$ & 0.195 & [200, 1] & [256, 128] \\
& Rate of Missing Posts & $5.970 \times 10^{-5}$ & 0.094 & [100, 1] & [256, 128] \\
\midrule
Sentence-BERT & Number of Posts & $6.728 \times 10^{-6}$ & 0.066 & [100, 1] & [1024, 512] \\
 & Rate of Censorship & $2.055 \times 10^{-5}$ & 0.276 & [100,1] & [512, 256] \\
& Rate of Missing Posts & $2.641 \times 10^{-5}$ & 0.071 & [50,1] & [512, 256] \\
\bottomrule
\end{tabular}}
\end{table}

We automated hyperparameter optimization using Optuna
\citep{akiba_optuna_2019}, selecting the best parameters based on
validation loss across 100 trials. For both hyperparameter tuning and
nuisance function estimation, models were trained for up to 10,000
epochs, with early stopping triggered if validation loss failed to
improve for five consecutive epochs. We used the Adam optimizer with a
batch size of 256 and a weight decay of $10^{-8}$ to prevent
overfitting. To stabilize training, we applied gradient clipping with
a maximum norm of 1.0. For the sake of completeness,
Table~\ref{tab:hyperparameters} reports the optimal hyperparameter
configurations for each outcome measure.

After selecting the optimal hyperparameters, we estimated the ATT for
each outcome using the DML procedure described in the previous
section. Specifically, we implemented two-fold cross-fitting to estimate
the nuisance components: the deconfounder $\boldf(\bR_i)$, the outcome
model $\mu_{T_i}(\boldf(\bR_i), \bm{Z}_i)$, and the propensity score
$\pi(\boldf(\bR_i), \bm{Z}_i)$.

The deconfounder and outcome model were trained using the selected
hyperparameters for up to 10,000 epochs, with early stopping triggered
if the validation loss did not improve for five consecutive
epochs. For propensity score estimation, we used a neural network with
spectral normalization \citep{gouk2021regularisation} to enforce
Lipschitz continuity. This network consisted of a two-layer MLP with
ReLU activation, using 128 hidden units in the first layer and 64 in
the second. The learning rate was fixed at $2 \times 10^{-5}$, and the model
was trained under the same early stopping criteria.

The ATT was then computed using \eqref{eq:tau.hat}, with standard
errors derived from the influence function given in \eqref{eq:psi}. To
account for within-user correlation, we clustered standard errors at
the user level. For consistency with the original study's text
matching approach, we estimated the ATT using both the full sample and
the matched sample from the original analysis. We also consider the
optimal combinations of the estimates from the two LLMs using the
optimal weights in Proposition~\ref{asymp_combined} so that we can have
more efficient results.

For comparison, we replicated the text matching procedure proposed by
\citet{roberts_adjusting_2020}, following their publicly available
replication materials. Specifically, we fit a structural topic model
with 100 topics, incorporating the censorship indicator (treatment) as
a covariate. From this model, we obtained both the estimated topic
proportions and treatment projections for each post.

Next, we applied coarsened exact matching (CEM) to pair censored posts
with uncensored ones based on four criteria: (1) topic proportions,
(2) treatment projection, (3) post date, and (4) prior censorship
history. This matching procedure yielded a final sample of 879 posts
from 628 users. Using the matched sample and weights generated by CEM,
we compared treatment and control posts across the three outcome
measures described earlier. As in the main analysis, standard errors
were clustered at the user level to account for within-user
correlation.

In addition, we replicated three representation-based approaches for
comparison. First, following \citet{veitch_adapting_2020,
  gui_causal_2023}, we fine-tuned a pre-trained language model to
construct text representations for causal inference. Specifically, we
used BERT embeddings \citep{devlin2019bert} as the initial text
representation and jointly optimized predictive losses for the
treatment assignment and outcome. Once the outcome model was
estimated, following \citet{gui_causal_2023}, we fit the propensity
score models conditional on the estimated outcome model under each
treatment arm, and finally estimated the treatment effect using the
influence function given in \eqref{eq:psi}. Because fine-tuning BERT
is computationally expensive, we adopted the default hyperparameters
from \citet{gui_causal_2023}, except for the learning rate, and
trained the model for 30 epochs. We set the learning rate to 0.001 to
accelerate convergence, so that the relatively small number of
training epochs would not materially affect the results.

Second, we used the same neural network architecture as in
Figure~\ref{tarnet}, but replaced the input representations with
alternative pre-trained language model representations. Specifically,
we used the multilingual sentence embedding model
\texttt{paraphrase-multilingual-MiniLM-L12-v2} from the
\texttt{sentence-transformers} library, which produces 384-dimensional
representations. We also used the Qwen3 embedding model with 0.6
billion parameters, which produces 1024-dimensional representations;
this model is based on a large language model fine-tuned for embedding
tasks such as information retrieval and semantic similarity. Finally,
we used the same pre-trained embedding representations from Qwen3 and
Sentence-BERT but used the DragonNet \citep{shi_adapting_2019}, which
added the treatment prediction head based on the shared representation
of the neural network in Figure~\ref{tarnet} while minimizing the loss
function in \eqref{loss_func2}.

Unlike the internal representations used in GPI, these embeddings are
not designed to exactly reconstruct the input text, which makes them
cheaper to compute but may discard information relevant to downstream
causal estimation. By comparing these results with the original GPI
estimates, we can assess whether the performance gains arise from
better internal representations. For each pre-trained embedding, we
fine-tuned using the same procedure as GPI and used the exact same
architecture for the sake of
comparison. Table~\ref{tab:hyperparameters} shows the selected
hyperparameter values.

\subsection{Additional Empirical Results}
\label{subsec:additional_results_text}

Table~\ref{tab:gpi_results} presents the results for the GPI
estimators based on LLaMA3, Gemma3, and their optimal combination,
while Table~\ref{tab:alternative_results} reports results using
alternative text representations and benchmark methods, including
Qwen3--0.6B embeddings, Sentence-BERT, fine-tuned-BERT+DragonNet
\citep{veitch_adapting_2020}, and text matching.

The fine-tuned BERT model with DragonNet proposed by
\citet{veitch_adapting_2020} produces better estimates, but these
estimates are still statistically significantly different from those
obtained using the GPI methods for the rate of censorship and the rate
of missing posts.

In addition to the findings summarized in Section~\ref{sec:confounder}
of the main text, we find that the relative efficiency of GPI and text
matching varies across outcomes. Although GPI yields smaller standard
errors for the number of posts after censorship, text matching yields
smaller standard errors for the rate of censorship and the rate of
missing posts. This pattern, however, should be interpreted with
caution. The text-matching estimator treats the topic-model
representations as fixed and therefore does not account for the
uncertainty involved in estimating these representations. As a result,
its reported standard errors may understate the true uncertainty. By
contrast, GPI incorporates the uncertainty from estimating the
deconfounder within a semiparametric estimation framework. At the same
time, GPI tunes the representation to optimize out-of-sample outcome
prediction, prioritizing features that explain outcome variation.

We also find that adding the treatment head changes the point
estimates for the rate of censorship. For the Qwen embedding, the
estimated effect is attenuated by 30\% in the full sample and by 50\%
in the matched sample. The differences are statistically significant
for both Qwen and Sentence-BERT. By contrast, DragonNet does not
materially change the results for the other two outcomes for either
pre-trained embedding model.

\begin{table}[t]
\centering
\caption{Estimated Average Treatment Effects for the Treated (ATT) using GPI with LLaMA3 (8B), Gemma3 (1B), and their optimal combination. Standard errors (reported in parentheses) are clustered at the user level. Results are shown for both the full sample and the matched sample.}
\label{tab:gpi_results}
\spacingset{1}
\begin{tabular}{lcccccc}
\toprule
& \multicolumn{2}{c}{\textbf{GPI {\footnotesize (LLaMA3-8B)}}} &
\multicolumn{2}{c}{\textbf{GPI {\footnotesize (Gemma3-1B)}}} &
\multicolumn{2}{c}{\textbf{GPI {\footnotesize (Combined)}}} \\
\textbf{Outcome} & Full & Matched & Full & Matched & Full & Matched \\
\midrule

Rate of censorship
& 0.012 & 0.018 & 0.011 & 0.015 & 0.011 & 0.017 \\
& (0.000) & (0.002) & (0.000) & (0.002) & (0.000) & (0.002) \\

Rate of missing posts
& 0.065 & 0.100 & 0.070 & 0.099 & 0.067 & 0.099 \\
& (0.003) & (0.022) & (0.003) & (0.023) & (0.003) & (0.021) \\

Number of posts
& -16.200 & -14.700 & -14.400 & -4.330 & -15.600 & -7.690 \\
& (2.600) & (21.900) & (2.690) & (21.300) & (2.560) & (21.100) \\

\bottomrule
\end{tabular}
\end{table}

\begin{table}[t]
\centering
\caption{Estimated Average Treatment Effects for the Treated (ATT) using alternative text representations and benchmark methods. Standard errors (reported in parentheses) are clustered at the user level. Results are shown for both the full sample and the matched sample whenever applicable.}
\label{tab:alternative_results}
\spacingset{1}
\resizebox{\textwidth}{!}{%
\begin{tabular}{lccccccccccc}
\toprule
& \multicolumn{2}{c}{\textbf{\shortstack{Qwen Embedding \\ + TarNet}}} &
  \multicolumn{2}{c}{\textbf{\shortstack{Qwen Embedding \\ + DragonNet}}} &
  \multicolumn{2}{c}{\textbf{\shortstack{Sentence-BERT \\ + TarNet}}} &
  \multicolumn{2}{c}{\textbf{\shortstack{Sentence-BERT \\ + DragonNet}}} &
  \multicolumn{2}{c}{\textbf{\shortstack{BERT (Fine-tuned) \\ + DragonNet}}} &
  \multicolumn{1}{c}{\textbf{Text Matching}} \\
\textbf{Outcome}
& Full & Matched
& Full & Matched
& Full & Matched
& Full & Matched
& Full & Matched
& Matched \\
\midrule

Rate of censorship
& 0.010 & 0.018
& 0.007  & 0.009
& -0.049 & -0.101
& 0.008  & 0.012
& 0.014 & -0.001
& 0.004 \\
& (0.000) & (0.005)
& (0.000) & (0.002)
& (0.027) & (0.312)
&  (0.000) & (0.002)
& (0.001) & (0.002)
& (0.001) \\

Rate of missing posts
& 0.053 & 0.077
& 0.058 & 0.091
& 0.049 & 0.082
& 0.064 & 0.069
& 0.057 & -0.030
& 0.050 \\
& (0.002) & (0.026)
& (0.003) & (0.022)
& (0.003) & (0.030)
& (0.003) & (0.026)
& (0.003) & (0.021)
& (0.016) \\

Number of posts
& -16.441 & 9.223
& -18.015 & 0.432
& -23.000 & -10.621
& -23.938 & -7.516
& -22.788 & 8.635
& 11.500 \\
& (2.313) & (23.648)
& (2.520) & (19.179)
& (2.520) & (23.545)
& (2.516) & (17.791)
& (3.806) & (21.758)
& (41.200) \\

\bottomrule
\end{tabular}}
\end{table}

\begin{table}[t]
\caption{Correlation between estimated influence functions from GPI
  using LLaMA3-8B and Gemma3-1B models. Results are reported for
  both the full and matched samples.}\label{if_corr}
\centering
\begin{tabular}{lcc}
\toprule
& \textbf{Full} & \textbf{Matched} \\
\midrule
Rate of censorship
  & 0.947 & 0.941 \\
Rate of missing posts
  & 0.747 & 0.663 \\
Number of Posts
  & 0.884 & 0.922 \\
\bottomrule
\end{tabular}
\end{table}

\subsection{Details of the Robustness Analysis}
\label{subsec:robustness}

To further assess the robustness of our approach, we examine how much
the point estimates change when we additionally adjust for textual
features that might act as a confounder.  Specifically, we use a set
of 60 keywords independently identified by \citet{fu2013assessing}: (i)
30 keywords with the highest relative frequency in censored posts
compared to uncensored ones, and (ii) 30 keywords most frequent among
self-censoring users relative to others (as reported in Appendix
Tables~1a~and~1b of \citealt{fu2013assessing}). Then, our candidate
confounder is the proportion of these keywords in each post.

We estimate both the GPI and text matching methods with and without
directly controlling these text features, and calculate the relative
absolute bias (i.e., the absolute change in point estimates divided by
the absolute value of the original point estimates). For text
matching, we additionally control for these textual features by using
the same structural topic model specification while incorporating the
60 keywords as controls in the coarsened exact matching (CEM). For
GPI, we calculate the standard error of the relative absolute bias
using the delta method, as shown below.  For text matching, since no
analytic variance formula is available, we estimate uncertainty using
the bootstrap with 200 replications though the validity of this
uncertainty estimate is not guaranteed \citep{abadie2016matching}.

Formally, let
$\tau_{\mathrm{long}}$ and $\tau_{\mathrm{short}}$ be the parameter of
interest obtained from GPI with and without controlling the text
features, respectively, and let $C_i$ be the text-based confounding
feature for post $i$.  Then, the relative absolute bias is defined as
$$
R = \frac{|\tau_{\mathrm{long}} - \tau_{\mathrm{short}}|}{|\tau_{\mathrm{short}}|}.
$$
Let $\psi_{\mathrm{long}}$ and $\psi_{\mathrm{short}}$ be the influence functions of $\tau_{\mathrm{long}}$
and $\tau_{\mathrm{short}}$, respectively. Because they are influence functions, we have
$$
\sqrt{N}
\begin{pmatrix}
\hat{\tau}_{\mathrm{long}} - \tau_{\mathrm{long}} \\
\hat{\tau}_{\mathrm{short}} - \tau_{\mathrm{short}}
\end{pmatrix}
=
\frac{1}{\sqrt{N}} \sum_{i=1}^n
\begin{pmatrix}
\psi_{\mathrm{long}}(\cD_i, C_i; \tau_{\mathrm{long}}, \boldf_{\mathrm{long}}, \mu_{1, \mathrm{long}}, \mu_{0,\mathrm{long}}, \pi_{\mathrm{long}}) \\
\psi_{\mathrm{short}}(\cD_i; \tau_{\mathrm{short}}, \boldf_{\mathrm{short}}, \mu_{1, \mathrm{short}}, \mu_{0,\mathrm{short}}, \pi_{\mathrm{short}})
\end{pmatrix}
+ o_P(1)
$$
where $\boldf_{\mathrm{long}}$, $\mu_{t,\mathrm{long}}$, and $\pi_{\mathrm{long}}$ are the nuisance functions estimated
when controlling the text features, and $\boldf_{\mathrm{short}}$, $\mu_{t,\mathrm{short}}$, and $\pi_{\mathrm{short}}$ are those estimated without controlling the text features.
By the multivariate central limit theorem, we have
$$
\sqrt{N}
\begin{pmatrix}
\hat{\tau}_{\mathrm{long}} - \tau_{\mathrm{long}} \\
\hat{\tau}_{\mathrm{short}} - \tau_{\mathrm{short}}
\end{pmatrix}
\xrightarrow{d}
\mathcal{N}\left(
\begin{pmatrix}
0 \\ 0
\end{pmatrix},
\begin{pmatrix} \sigma_{\mathrm{long}}^2 & \sigma_{\mathrm{long, short}} \\
\sigma_{\mathrm{long, short}} & \sigma_{\mathrm{short}}^2
\end{pmatrix}
\right)
$$
where $\sigma_{\mathrm{long}}^2 = \E[\psi_{\mathrm{long}}^2]$,
$\sigma_{\mathrm{short}}^2 = \E[\psi_{\mathrm{short}}^2]$, and
$\sigma_{\mathrm{long, short}} = \E[\psi_{\mathrm{long}} \psi_{\mathrm{short}}]$.
Then, under the assumption that $\tau_{\mathrm{short}} \ne 0$ and $\tau_{\mathrm{long}} - \tau_{\mathrm{short}} \ne 0$,
by the delta method, we have
$$\sqrt{n}(\hat{R} - R) \xrightarrow{d} \mathcal{N}(0, V) \quad
\text{where} \quad V = \nabla g^\top
\begin{pmatrix} \sigma_{\mathrm{long}}^2 & \sigma_{\mathrm{long, short}} \\
\sigma_{\mathrm{long, short}} & \sigma_{\mathrm{short}}^2
\end{pmatrix}
\nabla g$$
with $g(x,y) = |x - y|/|y|$ and $\nabla g = (\partial g/\partial x, \partial g/\partial y)^\top$ evaluated at $(\tau_{\mathrm{long}}, \tau_{\mathrm{short}})$.

Table~\ref{correlation_app_gpi} reports the results for the GPI
estimators, while Table~\ref{correlation_app_alternatives} reports the
results using alternative representations and benchmark methods. We
find that the GPI methods achieve relatively small absolute bias
though the results based on Qwen embedding are similar.  The estimates
based on Sentence-BERT with TarNet are also similar, though we see
that those estimates are different from the ones from GPI as reported
in Table~\ref{tab:alternative_results}.  The text matching is the
least robust to the inclusion of additional confounding features,
suggesting that in high-dimensional settings matching methods have a
difficulty in adjusting for confounding features.

\begin{table}[!t]
\centering
\caption{Relative absolute bias for the GPI estimators. Relative
  absolute bias is defined as the absolute change in point estimates
  from including a text-based confounder, relative to the original
  estimates. For the text-based confounder, we use the proportion of
  60 keywords related to censorship or self-censorship. Standard
  errors, reported in parentheses, are calculated by the delta
  method. ``Full'' refers to estimates using the entire sample, while
  ``Matched'' refers to estimates based on the matched sample.}
\label{correlation_app_gpi}
\spacingset{1}
\begin{tabular}{lcccc}
\toprule
& \multicolumn{2}{c}{\textbf{GPI {\footnotesize (LLaMA3-8B)}}}
& \multicolumn{2}{c}{\textbf{GPI {\footnotesize (Gemma3-1B)}}} \\
\textbf{Outcome} & Full & Matched & Full & Matched \\
\midrule
Rate of censorship & 0.007 & 0.196 & 0.037 & 0.132 \\
                   & (0.003) & (0.140) & (0.033) & (0.097) \\
Rate of missing posts & 0.040 & 0.279 & 0.124 & 0.142 \\
                      & (0.049) & (0.346) & (0.042) & (0.212) \\
Number of posts & 0.084 & 0.242 & 0.062 & 0.370 \\
                & (0.068) & (0.682) & (0.067) & (3.638) \\
\bottomrule
\end{tabular}
\end{table}

\begin{table}[!t]
\centering
\caption{Relative absolute bias for embedding-based estimators and
  text matching. Relative absolute bias is defined as the absolute
  change in point estimates from including a text-based confounder,
  relative to the original estimates. For the text-based confounder,
  we use the proportion of 60 keywords related to censorship or
  self-censorship. Standard errors, reported in parentheses, are
  calculated by the delta method for embedding-based estimators and by
  bootstrap for text matching, as an analytic formula is not available
  for text matching. ``Full'' refers to estimates using the entire
  sample, while ``Matched'' refers to estimates based on the matched
  sample.}
\label{correlation_app_alternatives}
\spacingset{1}
\resizebox{\textwidth}{!}{%
\begin{tabular}{lccccccccccc}
\toprule
& \multicolumn{2}{c}{\textbf{\shortstack{Qwen Embedding \\ + TarNet}}}
& \multicolumn{2}{c}{\textbf{\shortstack{Qwen Embedding \\ + DragonNet}}}
& \multicolumn{2}{c}{\textbf{\shortstack{Sentence-BERT \\ + TarNet}}}
& \multicolumn{2}{c}{\textbf{\shortstack{Sentence-BERT \\ + DragonNet}}}
& \multicolumn{2}{c}{\textbf{\shortstack{BERT (Fine-tuned) \\ + DragonNet}}}
& \multicolumn{1}{c}{\textbf{Text matching}} \\
\textbf{Outcome} & Full & Matched & Full & Matched & Full & Matched & Full & Matched & Full & Matched & Matched \\
\midrule
Rate of censorship & 0.096 & 0.095 & 0.045 & 0.083 & 0.110 & 0.806 & 1.403 & 6.906 & 0.133 & 1.093 & 0.202 \\
& (0.021) & (0.189) & (0.020) & (0.090) & (1.362) & (0.505) & (2.770) & (40.592) & (0.017) & (0.191) & (0.277) \\
Rate of missing posts & 0.008 & 0.030 & 0.086 & 0.072 & 0.037 & 4.086 & 0.032 & 0.236 & 0.082 & 2.016 & 0.627  \\
& (0.021) & (0.177) & (0.019) & (0.096) & (0.322) & (8.079) & (0.022) & (0.231) & (0.042) & (1.602) & (3.416) \\
Number of posts & 0.095 & 0.467 & 0.047 & 5.957 & 0.022 & 0.068 & 0.089 & 0.032 & 0.799 & 0.900 & 5.350 \\
& (0.039) & (2.291) & (0.045) & (259.396) & (0.021) & (0.476) & (0.032) & (1.934) & (0.339) & (4.429) & (33.794) \\
\bottomrule
\end{tabular}}
\end{table}

\section{Image as Treatment}\label{sec:ap_image}

In this section, we provide a formal theoretical justification of our
methodology introduced in Section~\ref{sec:image} of the main text and
describe its implementation details.

\subsection{Theoretical Properties}
\label{subsec:theory_image}

Our identification result is nearly identical to that of
\citet{imai2024causal}. The key difference is that we focus on image rather than text.
We also derive testable implications of our key identification condition (Assumption~\ref{separability}).
As in the previous section, we define the causal estimand of
interest, establish its nonparametric identification, and then develop
an appropriate estimation strategy.

\subsubsection{Assumptions and Identification}

Suppose we observe a sample of $N$ i.i.d. units $i = 1, \ldots, N$
from a population of interest. For each unit $i$, we observe an
unstructured treatment object
$\bX_i \in \mathcal{X} \subset \mathbb{R}^r$ (e.g., images or texts)
and an observed outcome $Y_i \in \cY \subset
\mathbb{R}$.
As in the
previous section, we assume that the treatment object $\bX_i$ is
generated by a deep generative model defined in
Definition~\ref{deep_use} and that the last layer of the deep
generative model is a deterministic function of the internal
representation $\bR_i$ (Assumption~\ref{det_dec}).

As before, we adopt the potential outcome framework.  We denote the
potential outcome under the treatment object $\bx \in \cX$ by
$Y_i(\bx)$, which is defined as the outcome that would be realized if
unit $i$ were to receive treatment object $\bx$.  The observed outcome
is denoted by $Y_i$ and is defined as the potential outcome under the
realized treatment object $\bX_i$, i.e., $Y_i = Y_i(\bX_i)$. This
notation assumes no interference between units and no hidden multiple
version of treatment objects \citep{rubi:90}. In our setup, since the
same annotator might code the same images multiple times, this
assumption implies the absence of carry-over effects. This is
formalized as follows:
\begin{assumption}[Consistency] \label{consistency} \spacingset{1} The potential
    outcome under the treatment object $\bx \in \cX$ is denoted by
    $Y_i(\bx)$ and equals the observed outcome $Y_i$ under the
    realized treatment object $\bX_i$:
    $$Y_i \ = \ Y_i(\bX_i).$$
\end{assumption}

Since we are working in the observational studies setting without
randomization, we rely on the standard ignorability assumption for
causal identification. Specifically, we assume that the assignment of
the treatment object is independent of the potential outcomes.  This
is satisfied in our application if the annotators do not select which
images to annotate. The assumption is formalized
as follows:
\begin{assumption}[Ignorability] \label{ignorability}
  \spacingset{1} The assignment of treatment object $\bX$ is independent of potential outcomes,
  $$Y_i(\bx) \ \indep \ \bX_i,$$
  where $0 < \P(\bX_i = \bx)< 1$ for all
  $i=1,\ldots,N$, $\bx \in \cX$.
\end{assumption}

We are interested in estimating the causal effect of a particular
feature that is a part of the treatment object. We assume that this
treatment feature $T$ is determined solely by the treatment object
$\bX$.
\begin{assumption}[Treatment Feature]\label{det_trt} \spacingset{1}
There exists a deterministic function $g_{T}: \cX \to \cT$ that maps a treatment object $\bX_i$ to a treatment feature of interest $T_i \in \cT$, i.e., $$T_i \ = \ g_T(\bX_i).$$
\end{assumption}
The assumption implies that the treatment variable is a function of
the treatment object and does not vary across respondents.

Next, we define confounding features, which represent all features of
$\bX$ other than the treatment feature $T$ that influence the outcome
$Y$.  These confounding features, denoted by $\bU_i$, are based on a
vector-valued deterministic function of $\bX$ and are denoted by
$\bU \in \cU$ where $\cU$ denotes their support. As in the previous
section, however, this confounding feature is not observed.
\begin{assumption}[Confounding Features]\label{confounder} \spacingset{1}
  There exists an unknown vector-valued deterministic function $\bg_{\bU}: \cX \to \cU$ that maps an unstructured object $\bX_i \in \cX$ to the confounding features $\bU_i \in \cU$, i.e.,
  $$\bU_i \ = \ \bg_{\bU}(\bX_i),$$
  where $\dim(\bU_i) \ll \dim(\bX_i)$.
\end{assumption}
We note that Assumptions~\ref{det_trt}--\ref{confounder} are identical to the ones presented in \citet{imai2024causal}.

Finally, we introduce our key identification assumption, which states
that it is possible to intervene the treatment feature without
changing the confounding features. Specifically, we assume that the
treatment feature cannot be represented as a deterministic function of
the confounding features. In addition, the confounding features should
not be a function of treatment feature either so that we only adjust
for pretreatment variables \citep{daoud_conceptualizing_2022}. We
formalize this assumption as follows.
\begin{assumption} {\sc (Separability of Treatment and Confounding Features)} \label{separability} \spacingset{1} The potential outcome is a function of the treatment feature of interest $t$ and another separate function of the confounding features $\bu$.  Specifically, for any given $\bx \in \cX$ and all $i$, we have:
$$
Y_i(\bm{x}) = Y_i(t, \bu)=Y_i(g_{T}(\bx), \bg_{\bU}(\bx)),
$$
where $t=g_{T}(\bx) \in \cT$ and $\bu=\bg_{\bU}(\bx) \in \cU$. In
addition, $g_T$ and $\bg_{\bU}$ are separable.  That is, there exists
no deterministic function $\tilde{g}_t: \cU \to \{0,1\}$, which
satisfies
$\mathbf{1}\{g_T(\bx)=t\}= \tilde g_t\bigl(\bg_{\bU}(\bx)\bigr)$ for
all $\bx \in \cX$ and any $t \in \cT$. Similarly, there exist no
deterministic functions $\bg^\prime: \cX \to \cX^\prime$ and
$\tilde{\bg}_{\bU}: \cT \times \cX^\prime \to \cU$ that depend
nontrivially on the first argument, which satisfy
$\bg_{\bU}(\bx) = \tilde{\bg}_{\bU} ( g_T(\bx), \bg^\prime(\bx))$ for
all $\bx \in \cX$ and
$\tilde{\bg}_{\bU} (t, \bg^\prime(\bx^\prime))\ne \tilde{\bg}_{\bU}
(t', \bg^\prime(\bx^\prime))$ for some $t \neq t'$ with $t,t'
\in \cT$ and some $\bx^\prime \in \cX$.
\end{assumption}
We formulate the separability assumption in terms of deterministic
functions rather than stochastic relationships because both the
treatment and confounding features are assumed to be deterministic
functions of the same unstructured data.  The first requirement of
separability states that the treatment feature should not be a
deterministic function of the confounding features, which implies the
overlap assumption (see Lemma~1 of \citealt{imai2024causal}). The second requirement states that the confounding
features should not be a function of the treatment feature, which
corresponds to the standard assumption that confounders are
pre-treatment variables.  We consider
$\bg_{\bU}(\bx) = \tilde{\bg}_{\bU} ( g_T(\bx), \bg^\prime(\bx))$
rather than $\bg_{\bU}(\bx) = \tilde{\bg}_{\bU} ( g_T(\bx))$ to allow
for the possibility that the confounding features $\bg_{\bU}$ may
depend on the unstructured object $\bX$ other than through the
treatment feature $g_T$ alone.

We show that this separability assumption implies the independence of
support, which has appeared as an operationalization of
disentanglement in the causal representation
learning literature \citep{wang2024desiderata}.
\begin{proposition}[Separability implies independent support]\label{indep_sup}
Under Assumption~\ref{separability}, for every \(t\in\cT\),
\[
    \mathrm{supp}(\bU_i \mid T_i=t)
    =
    \mathrm{supp}(\bU_i),
\]
where \(\mathrm{supp}(\cdot)\) denotes the support of the corresponding
random variable.
\end{proposition}
The proof is given in Appendix~\ref{proof_indep_sup}. This result is
analogous to Theorem~9 of \citet{wang2024desiderata} except that our
result concerns deterministic functions, while theirs is formulated
for stochastic mappings.

Under this setup, we are interested in estimating the following average causal
effect of the treatment feature (ATE) while adjusting for the confounding
features,
\begin{equation}
      \tau \ :=  \ \E[Y_i(1, \bU_i) - Y_i(0, \bU_i)]. \label{eq:curve}
\end{equation}

\begin{figure}[t]
\centering \spacingset{1}
\includegraphics[width=0.6\textwidth]{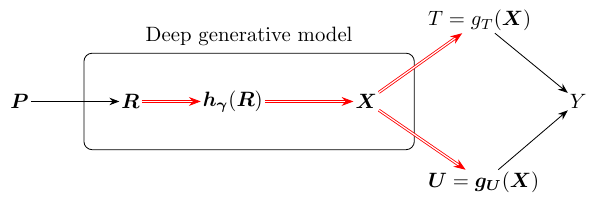}
\caption{Directed Acyclic Graph of the Assumed Data Generating Process
  when the Treatment is an Unstructured Object. A treatment object
  $\bX$ (e.g., image) is generated using a deep generative model
  (rectangle) with a prompt $\bP$, producing an internal
  representation $\bR$ that generates $\bX$ through a deterministic
  function $\bh_{\bgamma}(\bR)$. The treatment object affects the
  outcome $Y$ through its treatment feature of interest $T$ and other
  confounding features $\bU$. Treatment object is assumed to be
  independent of potential outcomes. An arrow with red double
  lines represents a deterministic causal relation while an arrow with
  a single line indicates a possibly stochastic relationship.}
\label{DAG2}
\end{figure}

Figure~\ref{DAG2} presents a DAG that summarizes the data generating
process and required identification assumptions (i.e.,
Assumptions~\ref{det_dec},~and~\ref{consistency}--\ref{separability}).
As in Figure~\ref{DAG1}, an arrow with double lines represents a
deterministic causal relationship, while an arrow with a single line
indicates a possibly stochastic causal relationship.

Given this setup, we establish the nonparametric identification of the
mean of the potential outcome.
\begin{proposition} {\sc (Nonparametric Identification of the Marginal Mean of Potential Outcome)}\label{iden_image} \spacingset{1}
  Under
  Assumptions~\ref{det_dec},~\ref{consistency}--\ref{separability},
  there exists a deconfounder function
  $\boldf: \cR \rightarrow \cQ \subset \R^{d_Q}$ with
  $d_Q = \dim(\cQ) \le d_R=\dim(\cR)$ that satisfies the following
  conditional mean independence relation: \begin{equation} \E[Y_i \mid
    \bR_i, T_i, \boldf(\bR_i)] = \E[Y_i \mid T_i,
    \boldf(\bR_i)]
  \label{eq:YindepRgivenTf}
\end{equation}
where $0<\P(T_i = t \mid \boldf(\bR_i)=\bq)<1$ for all $t \in \cT$
and $\bq \in \cQ$. In addition, the treatment feature and a
deconfounder are separable.  By adjusting for such a deconfounder, we
can uniquely and nonparametrically identify the marginal mean of the
potential outcome under the treatment condition $T_i = t$ for
$t\in \cT$ as:
  \begin{align*}
    \E[Y_i(t, \bU_i)] \ = \ \int_{\cR} \E[Y_i \mid T_i = t,
    \boldf(\bR_i)] dF(\bR_i).
  \end{align*}
\end{proposition}
We omit the proof since it is similar to that of Proposition~1 of
\citet{imai2024causal}.  We also do not present the estimation
procedure details, which are almost identical to the previous
application.

\subsection{Implementation Details for the Empirical Application}

We analyze the UCLA Protest Image Dataset \citep{won2017protest},
which contains 11,659 protest images in total. We focus on a subset of
9,316 images with human annotations for multiple attributes collected
via Amazon Mechanical Turk (MTurk). For objective binary attributes,
such as the presence of protest activities or specific visual scene
characteristics, each image was independently annotated by two MTurk
workers, with disagreements resolved by a third annotator.  In
contrast, perceived violence was measured through pairwise comparisons
because it is inherently subjective and continuous. Specifically, the
original study randomly sampled 58,295 image pairs, with each image
appearing in approximately 10 comparisons. For each pair, 10 MTurk
workers selected the image they perceived as more violent.

The authors
then used the Bradley-Terry model to estimate continuous image-level
perceived violence scores, assigning each image a real-valued score
between 0 and 1. See \citet{won2017protest} for additional details
about the dataset.

In our analysis, we treat nighttime (binary) as the treatment and
perceived violence (continuous, ranging from 0 to 1) as the
outcome. We expect the true causal effect of nighttime on perceived
violence to be small because an image can, in principle, be
transformed from nighttime to daytime without changing the underlying
level of violence, although such a transformation may still influence
how violent the image is perceived to be. Consequently, after
adjusting for other image features, the estimated effect of nighttime
on perceived violence should be substantially attenuated. This is
because nighttime can influence the outcome only through annotators'
perceptions rather than through any change in the actual level of
violence depicted in the image.

We implemented our proposed methodology as follows. We first
regenerated the protest images using Stable Diffusion
versions~1.5~and~2.1 \citep{rombach_high-resolution_2022}, then
extracted the internal representation from the final layer of the
diffusion model, immediately prior to the decoder of variational
autoencoder. Consistent with Assumption~\ref{det_dec}, we verified
that this decoder is a deterministic function of the internal
representation. Because image dimensions vary across the dataset, we
padded all images to $512 \times 512$ before passing them through the
Stable Diffusion models. The resulting internal representation has
dimensionality of $16{,}384 = 64 \times 64 \times 4$, which is
substantially smaller than the original image dimensionality of
$786{,}432 = 512 \times 512 \times 3$.

We fitted the neural network model that includes the internal
representation $\bR$, the treatment feature of interest (nighttime),
and the outcome variable (perceived violence). Unlike the text-based
application, we did not apply a pooling operation to reduce
dimensionality. Instead, we used a convolutional neural network (CNN)
as an encoder to further reduce the dimensionality of the internal
representation and then applied the same downstream neural network
architecture as in the previous application.  For the CNN layers, we
used kernel size 3, stride 1, padding 1, and pooling kernel size 2 for
the two versions of Stable Diffusion. The tuning procedure followed
the same protocol as before, with additional tuning for the CNN
layers. Specifically, the hyperparameter search space for the CNN
layers included:
\begin{itemize}
    \item CNN channel architecture: either $[32,64]$, $[64,128]$, or $[32,64,128]$;
    \item CNN dropout: from $0.0$ to $0.3$;
    \item CNN batch normalization: either true or false.
\end{itemize}
Table~\ref{tab:hyperparameters_image} reports the selected optimal
hyperparameters using Optuna.

\begin{table}[t]
\centering
\caption{Hyperparameter values selected using Optuna for the image application.}
\label{tab:hyperparameters_image}
\spacingset{1}
\resizebox{\textwidth}{!}{%
\begin{tabular}{llcccccc}
\toprule
\textbf{Model} & \textbf{Learning Rate} & \textbf{Dropout} & \textbf{CNN Channels} & \textbf{CNN Dropout} & \textbf{CNN Batch Norm} & \textbf{Outcome Model} & \textbf{Deconfounder} \\
\midrule
Stable Diffusion v1.5
& $8.996 \times 10^{-5}$
& $0.276$
& $[32,64,128]$
& $2.903 \times 10^{-4}$
& False
& $[200,1]$
& $[512,256]$ \\

Stable Diffusion v2.1
& $6.109 \times 10^{-5}$
& $0.088$
& $[32,64]$
& $0.008$
& True
& $[50,1]$
& $[256,128]$ \\
\bottomrule
\end{tabular}}
\end{table}

To estimate the average effect of nighttime scene, we applied the
methodology described earlier, using two-fold cross-fitting in
combination with the selected hyperparameters to estimate both the
deconfounder and outcome models.  All optimization settings, including
the choice of optimizer, batch size, number of training epochs,
gradient clipping, and early stopping, were held consistent with the
prior application.  We also combined the estimates from the two
different generative models (Stable Diffusion versions~1.5~and~2.1) by
the optimal linear combination method discussed earlier (see
Proposition~\ref{asymp_combined}).

\section{Structural Model of Texts}\label{sec:ap_text}

In this section, we provide a formal theoretical justification for the
methodology introduced in Section~\ref{sec:text} of the main
text. This application is intended to demonstrate how our GPI
framework can be integrated into an existing structural
model. Specifically, we tailor the proposed methodology to the
experimental setting of \citet{blumenau_variable_2022}, which focuses
on estimating the latent persuasiveness of political rhetoric. Unlike
the previous two applications, this analysis employs a model-based
inferential approach. We begin by outlining the experimental design,
then introduce our proposed semiparametric model, and finally detail
the implementation of the empirical analysis.

\subsection{Experimental Design}
\label{subsec:design}

To estimate the latent persuasiveness of political rhetoric,
\citet{blumenau_variable_2022} constructed a set of hypothetical
political arguments and randomly assigned pairs of these arguments to
survey respondents. The authors designed a total of 336 distinct
arguments, which varied systematically across the following 12 policy issues in
contemporary British politics.
\begin{itemize}
\item Building a third runway at Heathrow
\item Closing large retail stores on Boxing Day
\item Extending the Right to Buy
\item Extension of surveillance powers in the UK
\item Fracking in the UK
\item Nationalization of the railways in the UK
\item Quotas for women on corporate boards
\item Reducing the legal restrictions on cannabis use
\item Reducing university tuition fees
\item Renewing Trident
\item Spending 0.7\% of GDP on overseas aid
\item Sugar tax in the UK
\end{itemize}
In addition, the authors used the following 14 rhetorical elements:
\begin{itemize}
\item Ad hominem
\item Appeal to authority
\item Appeal to fairness
\item Appeal to history
\item Appeal to national greatness
\item Appeal to populism
\item Common sense
\item Cost and benefit
\item Country comparison
\item Crisis
\item Metaphor
\item Morality
\item Public opinion
\item Side effects of policy
\end{itemize}
Lastly, the authors generated two versions of each argument,
corresponding to the ``for'' and ``against'' positions.

Once the set of arguments was generated, they were randomly assigned
to respondents. In the experiment, each respondent was presented with
a pair of arguments on the same topic---one in favor and one against a
given policy issue---that differed in their rhetorical
elements. Respondents were asked to indicate which argument they found
more persuasive, or whether they found both equally persuasive. A
total of 3,317 respondents participated in the study, each evaluating
four randomly selected issue pairs, resulting in 13,268 pairwise
observations.

\begin{table}[t]
	\caption{Two examples of political rhetorics about the policy
      issue regarding ``Building a third runway at Heathrow.''}
	  \vspace{0.2cm}
    \centering \spacingset{1}
    \begin{tabular}{|p{15cm}|}
       \hline
       \textbf{Appeal to authority / For} \\
        The Airports Commission, an independent body established to study the issue, have argued that expanding Heathrow is "the most effective option to address the UK’s aviation capacity challenge"\\
        \hline
      \textbf{Appeal to history / Against} \\
        History show us that most large infrastructure projects do not lead to significant economic growth, which suggests that the expansion of Heathrow will fail to pay for itself.\\
        \hline
    \end{tabular}
    \label{actual_arg}
\end{table}

Our goal is to estimate the latent persuasiveness of each rhetorical
element. The key methodological challenge is that, although argument
pairs are randomly assigned to respondents, the rhetorical elements
themselves may still be correlated with other features of the
texts. Table~\ref{actual_arg} provides two example arguments on the
same policy issue (``Building a third runway at Heathrow'') used in
the experiment. While both address the same topic, they differ in
multiple dimensions beyond rhetorical style or stance, highlighting
the potential for confounding.

To address this issue, \citet{blumenau_variable_2022} identified seven
additional textual features---argument length, readability, positive
and negative tone, overall emotional language, fact-based language,
and whether the argument was sourced from parliamentary speeches in
Hansard or authored by the researchers---and included them as
additional covariates in their robustness check. However, there is no
theoretical guarantee that this list of covariates fully captures all
relevant sources of variation, leaving open the possibility of
residual confounding.

\subsection{Model of Latent Persuasiveness}
\label{subsec:model_persuasion}

We model the persuasiveness of the political rhetoric in the way
similar to the original analysis of \citet{blumenau_variable_2022}, but
instead of a parametric model, we use a semiparametric model by
leveraging the internal representation of the deep generative model to
adjust for the argument-level confounders. Suppose that we have $J$
arguments indexed by $j = 1, \cdots, J$ and $N$ respondents indexed by
$i = 1, \cdots, N$. Let $\bX_{j}$ denote the text of argument $j$. We
regenerate this text using a deep generative model as defined in
Definition~\ref{deep_use} and obtain its internal representation
$\bR_{j}$. As before, we assume that the last layer of the deep
generative model is a deterministic function of the internal
representation $\bR_{j}$ (Assumption~\ref{det_dec}).

Each argument text $\bX_{j}$ contains three features of interest: a
rhetorical element $T_{j} \in \{1, \cdots, 14\}$, a side of the
argument $S_{j} \in \{1,2\}$ (for or against), and a policy issue
$P_{j} \in \{1, \cdots, 12\}$. All features of interest are solely
based on the argument text $\bX_{j}$, and thus can be written as
deterministic functions of the internal representation $\bR_{j}$.

Each respondent $i$ is presented with a pair of arguments
$(\bX_{j}, \bX_{j'})$ on the same topic $P_j = P_{j'}$ but with the
opposite sides $S_j \ne S_{j'}$ and then asked to answer the question
about persuasiveness. Let $Y_{ijj'}$ denote the outcome
variable, representing respondent $i$'s answer to this question. Then,
we can define the outcome as follows:
\begin{align*}
Y_{ijj'} = \begin{cases}
0 & \text{if respondent $i$ answers that argument $j$ is more
  persuasive than argument $j'$} \\
1 & \text{if respondent $i$ answers that both arguments are equally persuasive} \\
2 & \text{if respondent $i$ answers that argument $j'$ is more
  persuasive than argument $j$}
\end{cases}
\end{align*}
We use $\mathcal{J}(i)$ to represent a set of argument pairs assigned
to respondent $i$.

In the original analysis, \citet{blumenau_variable_2022} employed a
variant of the Bradley–Terry model \citep{bradley1952rank} to estimate
the latent persuasiveness of rhetorical elements. The model assumes
the following data-generating process,
\begin{align}
\log \left[ \frac{\P(Y_{ijj'} \leq y )}{\P(Y_{ijj'} > y)} \right] &= \delta_k + (\alpha_{P_j S_j} + \beta_{T_j} + \gamma_{j}) - (\alpha_{P_{j’} S_{j’}} + \beta_{T_{j’}} + \gamma_{j’}), \label{eq:original}
\end{align}
where $\beta_{t} \sim \mathcal{N}(0, \omega)$ and
$\gamma_j \sim \mathcal{N}(0, \sigma_{T_j})$. Here, $\delta_k$
denotes the intercept for response category $y$,
$\alpha_{ps}$ represents the fixed effect associated with policy issue
$P_j=p$ and argument side $S_j=s$, $\beta_{t}$ captures the average
effect of rhetorical element $T_j=t$, and $\gamma_{j}$ denotes the
random effect for argument $j$ with variance $\sigma_{T_j}$, which is
allowed to vary across rhetorical elements.

Importantly, this model adjusts only for policy issue and argument
side, and therefore does not account for other potentially confounding
features of the text. In addition, the model assumes independent
normal prior distributions for both $\beta_t$ and $\gamma_j$. Neither
assumption is implied by the experimental design. Our goal is to apply
GPI in order to relax these restrictive assumptions.

Specifically, we estimate the persuasiveness of rhetorical element
$T_{j}$ in argument $j$ while adjusting for confounding text features
$\bX_{j}$. As before, let $\bU_{j}$ denote the confounding features
of argument $j$ that are not captured by the rhetorical element
$T_{j}$. We assume that
the confounding features $\bU_{j}$ are a deterministic function of the
argument text $\bX_{j}$, i.e., $\bU_{j} = \bg_{\bU}(\bX_{j})$, where
$\bg_{\bU}$ is an unknown vector-valued function
(Assumption~\ref{confounder}). These confounding features are further
assumed to be separable from the rhetorical element $T_{j}$
(Assumption~\ref{separability}).

We model the outcome by adjusting for all the other relevant features
in the texts. Specifically, we assume the following semiparametric
structural model:
\begin{align}
&\log \left[ \frac{\P(Y_{ijj'} \leq y )}{\P(Y_{ijj'} > y)} \right] = \delta_y + \beta_{T_j} - \beta_{T_{j'}} +  h(\bR_{j}) - h(\bR_{j'}) \label{eq:structural_infeasible}
\end{align}
where we set $\sum_{j = 1}^J \beta_j = 0$ and
$\sum_{j = 1}^J h(\bR_{j}) = 0$ for identification, $\beta_{T_j}$
represents the persuasiveness of rhetorical element $T_j$, and
$h(\bR_{j})$ is the effect of confounding features. This model is
semiparametric because we do not restrict the functional form of
$h(\bR_{j})$ while assuming that the persuasiveness of each rhetorical
element $T_j = t \in \{1, \cdots, 14\}$ additively enters the model on
the log-odds scale. This additive specification is necessary because
the number of unique arguments is limited (i.e., $J=336$), making the
estimation of high-dimensional interactions among textual features
statistically unstable in this relatively small sample.

To estimate this model, we propose fitting the model with neural network architecture. Specifically, we use the following
semiparametric structural model for estimation:
\begin{align}
  \log \left[ \frac{\P(Y_{ijj'} \leq y )}{\P(Y_{ijj'} > y)} \right] &=
  \delta_y + \beta_{T_j} - \beta_{T_{j'}} +  h(\bR_{j}; \btheta) - h(\bR_{j'}; \btheta). \label{structural}
\end{align}
where $\btheta$ is the parameters of the persuasiveness of
confounding features $h(\bR_{j}; \btheta)$, and
$\sum_{j = 1}^{14} \beta_j = 0$ and $\sum_{j = 1}^J h(\bR_{j}; \bm\theta) = 0$.  This structural model is feasible because we observe
the internal representation $\bR_{j}$ and confounding feature $\bU_j$ is a deterministic function of internal representations.

We model $h(\bR_{j}; \btheta)$ as a neural network function of
$\bR_{j}$ using the standard feed-forward neural network
architecture. We minimize the following loss function to estimate the
parameters:
\begin{equation}
  \begin{aligned}
  \{ \hat{\bm\beta}, \hat{\btheta}, \hat{\bm \delta}\} &=
  \argmin_{\bm\beta,  \btheta, \bm{\delta}} - \sum_{i=1}^N
  \sum_{(j,j') \in \mathcal{J}(i)} \sum_{y = 0}^1\Bigl[ \mathbf{1}\{Y_{ijj'} \leq y\}\log \sigma(\delta_y + \beta_{jj'} + h_{jj'}) \\
  & \hspace{1.45in} + \mathbf{1}\{ Y_{ijj'} > y  \} \log \{ 1- \sigma(\delta_y + \beta_{jj'} + h_{jj'})\}\Bigr] \\
  &\text{s.t.} \quad \sum_{j = 1}^{14} \beta_j = 0, \qquad  \sum_{j = 1}^J h(\bR_{j}; \btheta) = 0,
  \end{aligned} \label{opt_latent}
\end{equation}
where $\bm\beta = (\beta_1, \cdots, \beta_{14})$,
$\bm\delta = (\delta_0, \delta_1)$, $\mathcal{J}(i)$ is the pair of
arguments that the respondent $i$ received,
$\beta_{jj'} = \beta_{T_j} - \beta_{T_{j'}}$,
$h_{jj'} = h(\bR_{j}; \btheta) - h(\bR_{j'}; \btheta)$, and
$\sigma(x) = 1/(1+\exp(-x))$ is the logistic function. We quantify the
uncertainty of the estimated persuasiveness parameters $\beta_t$ for
$t \in \{1, \cdots, 14\}$ by estimating standard errors via the
bootstrap and constructing confidence intervals using the normal
approximation.

\subsection{Implementation Details of the Empirical Application}
\label{subsec:implementation_persuasion}

To implement our proposed methodology, we first regenerated all 336
arguments using the three different LLMs: (1) LLaMA3 with eight
billion parameters, (2) LLaMA3.3 with seventy billion parameters, and
(3) Gemma3 with one billion parameters. Similar to the first
application, we extracted the internal representation of each
argument. The dimensions of internal representation are
different across models: 4096 for LLaMA3--8B, 8192 for LLaMA3.3--70B,
and 1152 for Gemma3--1B.

Even though the dimensions are different, since all internal
representations deterministically reproduce each argument, they should
contain all the information that are needed to reproduce the texts and
should yield similar estimates. More generally, the efficiency of GPI
estimators is governed by two competing factors: (i) the
dimensionality of the internal representation and (ii) the
informativeness of that representation. More capable foundation models
(e.g., LLaMA) often produce higher-dimensional representations, which
can reduce statistical efficiency by increasing estimation
variance. At the same time, these representations may contain richer
information that improves the prediction of outcomes and nuisance
functions. The resulting trade-off is therefore application-specific:
in some settings, the additional information outweighs the cost of
higher dimensionality, whereas in others it does not.

We fine-tuned the feed-forward neural networks within the
semiparametric structural model (see \eqref{structural}).  We
implemented the constraint in the optimization problem given in
\eqref{opt_latent} by demeaning the estimated outcome model, ensuring
its average is zero.
For fine-tuning, we employed the same hyperparameter search space and
procedure as in the previous two applications, except for the neural
network architecture, and optimized the hyperparameters using
Optuna. For the architecture, because the feed-forward network
$\tilde h$ models both the deconfounder $\boldf$ and the partially
linear component $h$ (i.e., $\tilde h = h \circ \boldf$), we
separately optimize the deconfounder and outcome-model components. For
both fine-tuning and the entire structural model fitting, we used the
Adam optimizer with batch size of 256,
and we applied gradient clipping with a maximum norm of
1.0. Table~\ref{tab:hyperparameters_structural} presents the optimal
hyperparameters selected by Optuna. We then fitted the entire
structural model using the selected hyperparameters, as described
above. Once we had fitted the outcome model, we estimated the latent
persuasiveness $\beta_{T_j}$ and its uncertainty by estimating the
standard error via the bootstrap with 200 bootstrap replications.

\begin{table}[t]
\centering
\caption{Optimal hyperparameter values selected for the semiparametric
  structural model.}
\label{tab:hyperparameters_structural}
\spacingset{1}
\resizebox{\textwidth}{!}{%
\begin{tabular}{lccccc}
\toprule
\textbf{Model} & $\dim(\bR)$ &  \textbf{Learning Rate} & \textbf{Dropout} & \textbf{Outcome Model} & \textbf{Deconfounder} \\
\midrule
LLaMA3 (8B) &  $4096$ & $6.073 \times 10^{-5}$ & 0.108 & [50, 1] & [512, 256]  \\
LLaMA3.3 (70B) & $8192$ & $2.897 \times 10^{-5}$ & 0.281 & [50, 1] & [1024, 512] \\
Gemma3 (1B) & $1152$ & $9.827 \times 10^{-5}$ & 0.280 & [50, 1] & [1024, 512]  \\
\bottomrule
\end{tabular}}
\end{table}

\begin{figure}[t]
\centering
\includegraphics[width=0.6\textwidth]{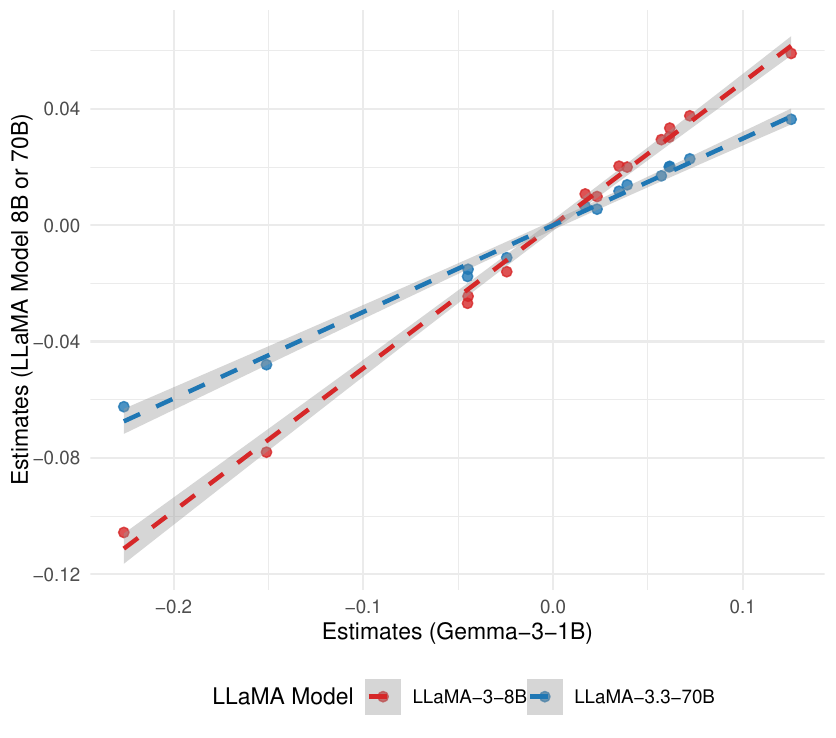}
\caption{The estimates of latent persuasiveness across three LLMs
  (LLaMA3 with eight billion parameters, LLaMA3.3 with seventy billion
  parameters, and Gemma3 with one billion parameters). The dashed
  lines represent the fitted regression lines and the gray area
  represents the 95\% confidence interval.  The results show that the
  estimates are highly correlated.}
  \label{fig:Text3_LLMpooledmean}
\end{figure}

Figure~\ref{fig:Text3_LLMpooledmean} shows that the estimates of latent
persuasiveness are consistent across the three LLMs.  The Pearson
correlations of the estimated latent persuasiveness of rhetorical
elements across models exceed 0.99 for all pairs of LLMs. These
results suggest that even though the internal representations are
different, the three LLMs yield similar estimates under the
deterministic decoding assumption
(Assumption~\ref{det_dec}).

\newpage
\section{Proofs}
\label{sec:proofs}

\subsection{Proof of Proposition~\ref{prop:identification}}\label{proof:prop_identification}
\begin{proof}
Under the definition of the deep generative model (see Definition~\ref{deep_use}) and Assumption~\ref{det_dec}, we can write $\bU_i = \boldf^*(\bR_i)$ with some deterministic function $\boldf^*$. Then,
\begin{align*}
  \E[Y_i(t)] &= \int_{\mathcal{Z} \times \cU} \E\bigl[Y_i(t) \mid \bm{Z}_i, \bU_i\bigr] dF(\bU_i, \bm{Z}_i)\\
  &= \int_{\mathcal{Z} \times \cU} \E\bigl[Y_i(t) \mid T_i = t, \bm{Z}_i, \bU_i\bigr] dF(\bU_i, \bm{Z}_i)\\
  &= \int_{\mathcal{Z} \times \cU} \E\bigl[Y_i \mid T_i = t, \bm{Z}_i, \bU_i\bigr] dF(\bU_i, \bm{Z}_i)\\
  &= \int_{\mathcal{Z} \times \cU} \E\bigl[Y_i \mid T_i = t, \bm{Z}_i, \boldf^*(\bR_i)\bigr] dF(\boldf^*(\bR_i), \bm{Z}_i)\\
  &= \int_{\mathcal{Z} \times \mathcal{R}} \E\bigl[Y_i \mid T_i = t, \bm{Z}_i, \boldf^*(\bR_i)\bigr] dF(\bR_i,\bm{Z}_i)
\end{align*}
where the second equality follows from Assumption~\ref{ignorability_confounder}, the third equality follows from Assumption~\ref{consistency_confounder}. Finally, suppose that there exists another function $\boldf(\bR_i)$ that satisfies $\E[Y_i \mid \bR_i, T_i = t, \bm{Z}_i, \boldf(\bR_i)] = \E[Y_i \mid T_i = t, \bm{Z}_i, \boldf(\bR_i)]$. Then,
\begin{align*}
&\int_{\mathcal{Z} \times \mathcal{R}} \E\bigl[Y_i \mid T_i = t, \bm{Z}_i, \boldf^*(\bR_i)\bigr] dF(\bR_i, \bm{Z}_i)\\
= & \int_{\mathcal{Z} \times  \mathcal{R}} \E\bigl[Y_i \mid T_i = t, \bm{Z}_i, \boldf(\bR_i), \bR_i \bigr] dF(\bR_i, \bm{Z}_i)\\
= & \int_{\mathcal{Z} \times \mathcal{R}} \E\bigl[Y_i \mid T_i = t, \bm{Z}_i, \boldf^*(\bR_i), \bR_i \bigr] dF(\bR_i, \bm{Z}_i)\\
= & \int_{\mathcal{Z} \times \mathcal{R}} \E\bigl[Y_i \mid T_i = t, \bm{Z}_i, \boldf^*(\bR_i)\bigr] dF(\bR_i, \bm{Z}_i)
\end{align*}
Thus, any function of $\bR_i$ that satisfies the aforementioned mean independence condition leads to the same identification formula.
\end{proof}

\subsection{Proof of Proposition~\ref{asymp_combined}}\label{proof:asymp_combined}
\begin{proof}
By Theorem \ref{asymp_text}, we know that each estimator $\hat\tau_j$ ($j = 1,2$) has the influence function representation:
\begin{align*}
\sqrt{N}(\hat\tau_j - \tau) = \frac{1}{\sqrt{N}} \sum_{i=1}^N \psi_j(\cD_{ij}) + o_P(1)
\end{align*}
where $\psi_j(\cD_{ij})$ is the influence function for the estimator $\hat\tau_j$ at the observation $\cD_{ij}$. Therefore, for any weight $\omega$, the estimator $\hat\tau = \omega \hat\tau_1 + (1-\omega) \hat\tau_2$ has the influence function representation:
\begin{align*}
\sqrt{N}(\hat\tau - \tau) = \frac{1}{\sqrt{N}} \sum_{i=1}^N \{ \omega \psi_1(\cD_{i1}) + (1-\omega) \psi_2(\cD_{i2}) \} + o_P(1)
\end{align*}
Therefore, by central limit theorem, for any $\omega$, we have
\begin{align*}
\sqrt{N}(\hat\tau - \tau) \xrightarrow{d} \mathcal{N}(0, V(\omega))
\end{align*}
where $V(\omega) = \E[( \omega \psi_1(\cD_{i1}) + (1-\omega) \psi_2(\cD_{i2}))^2]$.

Then, since $V(\omega)$ is a quadratic function of $\omega$, we can
minimize the asymptotic variance $V(\omega)$ with respect to $\omega$
as follows,
\begin{align*}
V(\omega) &= \omega^2 \E[\psi_1(\cD_{i1})^2] + (1-\omega)^2 \E[\psi_2(\cD_{i2})^2] + 2\omega(1-\omega) \E[\psi_1(\cD_{i1})\psi_2(\cD_{i2})]\\
&= \{ \E[\psi_1(\cD_{i1})^2] + \E[\psi_2(\cD_{i2})^2] - 2\E[\psi_1(\cD_{i1})\psi_2(\cD_{i2})] \} \omega^2\\
&\quad\quad + 2\{ \E[\psi_1(\cD_{i1})\psi_2(\cD_{i2})] - \E[\psi_2(\cD_{i2})^2] \} \omega + \E[\psi_2(\cD_{i2})^2].
\end{align*}
As it is a quadratic function of $\omega$, it is minimized at
\begin{align*}
\omega^* = \frac{\E[\psi_2(\cD_{i2})^2] - \E[\psi_1(\cD_{i1})\psi_2(\cD_{i2})]}{\E[\psi_1(\cD_{i1})^2] + \E[\psi_2(\cD_{i2})^2] - 2\E[\psi_1(\cD_{i1})\psi_2(\cD_{i2})]}
\end{align*}
\end{proof}

\subsection{Proof of Proposition~\ref{indep_sup}}\label{proof_indep_sup}
\begin{proof}
We prove it by contradiction. Suppose, for contradiction, that
\begin{align*}
\mathcal S_{TU}
\neq
\cT \times \cU.
\end{align*}
where $ S_{TU}
=
\{(g_T(\bx),{\bg}_{\bU}(\bx)):\bm x\in\mathcal{X}\}$ is the joint support of treatment feature and confounding feature.
This means that there exists a pair $(t^\dagger,\bm u^\dagger)
\in
\cT \times \cU
$
such that
$
(t^\dagger,\bm u^\dagger)
\notin
\mathcal S_{TU}
$.
Because $\bm u^\dagger \in \cU$, there exists some $\bx_0 \in \mathcal X$ satisfying $\bg_{\bU}(\bx_0)=\bm u^\dagger$ and $t_0 := g_T(\bx_0) \neq t^\dagger$. Now, we consider $\bm g'$ and $\tilde{\bm g}_{\bm U}$ as $\bm g'(\bm x)=\bm g_{\bm U}(\bm x)$
and
\begin{align*}
     \tilde{\bm g}_{\bm U}(t,\bm u) =
     \begin{cases}
     \bm u \quad \text{if } (t,\bm u) \neq (t^\dagger, \bm u^\dagger)\\
     \bm c \quad \text{if } (t,\bm u) = (t^\dagger,\bm u^\dagger)
     \end{cases}
\end{align*}
where $\bm c \neq \bm u^\dagger$. This is well-defined because $(t^\dagger,\bm u^\dagger)\notin\mathcal S_{TU}$ (so the value of $\tilde{\bm g}_{\bm U}$ at this point is unconstrained by the requirement that it reproduce the observed data; the independence of support means that we can change $T$ and $\bm U$ separately). Now, by definition of $\tilde{\bm g}_{\bm U}$,
\[
\tilde{\bm g}_{\bm U}(t_0, \bm g'(\bm x_0))
=
\tilde{\bm g}_{\bm U}(t_0,\bm u^\dagger)
=
\bm u^\dagger,
\]
while
\[
\tilde{\bm g}_{\bm U}(t^\dagger,\bm g'(\bm x_0))
=
\tilde{\bm g}_{\bm U}(t^\dagger,\bm u^\dagger) = \bm c
\neq
\bm u^\dagger.
\]
Thus, $\tilde{\bm g}_{\bm U}(t,\bm g'(\bm x_0))$ changes with $t$ while holding $g'(\bm x_0)$ fixed. Therefore, $\bm U$ can be represented as a nontrivial deterministic function of
$T$ and another feature $g'(\bm x)$, contradicting the separability assumption. Hence no such unattainable pair $(t^\dagger,\bm u^\dagger)$ can exist under separability.
Therefore, $\mathcal S_{TU}
=
\cT \times \cU
$.
\end{proof}

\end{document}